\documentclass[10pt]{article} 
\usepackage[preprint]{tmlr}

\usepackage{amsmath,amsfonts,bm}

\def\eqref#1{equation~\ref{#1}}

\def\1{\bm{1}}

\DeclareMathAlphabet{\mathsfit}{\encodingdefault}{\sfdefault}{m}{sl}
\SetMathAlphabet{\mathsfit}{bold}{\encodingdefault}{\sfdefault}{bx}{n}

\usepackage{hyperref}
\usepackage{url}

\usepackage{amsmath,amsthm,amssymb}

\usepackage{mathtools}
\usepackage{subcaption}
\usepackage{enumitem}
\usepackage{algorithm}
\usepackage{algpseudocode}
\usepackage{booktabs}
\usepackage{multirow}
\usepackage{bbm}
\usepackage{adjustbox}

\numberwithin{equation}{section}
\newtheorem{theorem}{Theorem}[section]
\newtheorem{proposition}[theorem]{Proposition}
\newtheorem{corollary}[theorem]{Corollary}
\newtheorem{lemma}[theorem]{Lemma}
\theoremstyle{assumption}

\theoremstyle{definition}

\theoremstyle{remark}

\title{CUBE: Contrastive Understanding by Balanced Experiments}

\author{\name Dongseok Kim\thanks{These authors contributed equally.} \email jkds5920@gachon.ac.kr \\
      \addr Department of Computer Engineering\\
      Gachon University
      \AND
      \name Hyoungsun Choi\footnotemark[1] \email hschoi@gachon.ac.kr \\
      \addr Department of Computer Engineering\\
      Gachon University
      \AND
      \name Mohamed Jismy Aashik Rasool\footnotemark[1] \email aashikrasool@gachon.ac.kr \\
      \addr Department of Computer Engineering\\
      Gachon University
      \AND
      \name Gisung Oh\thanks{Corresponding author.} \email eustia@gachon.ac.kr \\
      \addr Department of Computer Engineering\\
      Gachon University}

\def\month{MM}  
\def\year{YYYY} 
\def\openreview{\url{https://openreview.net/forum?id=XXXX}} 

\begin{document}

\maketitle

\begin{abstract}
We introduce CUBE, Contrastive Understanding by Balanced Experiments, to estimate main effects and pairwise Banzhaf interactions over a fixed two-state probe space using shared balanced queries. Each design cancels constant and odd-order contamination from pairwise estimates, while randomization ensures unbiasedness and explicit alias probabilities yield exact mean squared errors without sparsity assumptions or order truncation. Matching risk bounds establish asymptotic minimax optimality within a specified class of nonadaptive, equal-weight, reversal-symmetric balanced contrast estimators, as admissible budgets grow while remaining small relative to the full probe space. Experiments reveal different interaction-specific risks despite identical output distributions and low-order effects, and different sign risks despite equal mean squared errors. Across evaluated tabular models, CUBE generally outperforms several sampling-based estimators, while Faith-Banzhaf-2 achieves the lowest mean error at the largest tested budgets.
\end{abstract}

\section{Introduction}
\label{sec:introduction}

Feature interactions describe how the effect of one input changes with the states of other inputs, supporting model inspection and debugging. Estimating these interactions from prediction queries becomes expensive as the number of features grows: even a two-state probe space contains $2^d$ configurations. Query-efficient explanation therefore depends on arranging a small set of evaluations that preserves information about the effects of interest.

A central difficulty is contamination between effects. A weak pairwise interaction may coexist with much larger main effects whose contributions fail to cancel in sampled queries. Higher-order components can also coincide with the target contrast on a reduced design. Such aliasing makes the accuracy of an explanation depend on both the response structure and the query arrangement. Factorial balance provides a way to eliminate selected sources of contamination exactly, while randomization controls the contribution of the remaining aliases.

We introduce CUBE, \emph{Contrastive Understanding by Balanced Experiments}, for estimating main effects and pairwise interactions over a fixed two-state probe space. CUBE combines distinct binary character columns, independent level permutations, and global reversal of every query configuration. Each experiment reuses the same evaluations across all target effects. The construction makes pairwise estimates invariant to arbitrary additive and odd-order response components, while random level permutations yield unbiased estimates for arbitrary finite responses. The feature states and background define the explanation target, and the admissible query budget determines the design size.

The resulting analysis links explanation error directly to alias collisions. Their probabilities depend only on the query budget and the symmetric difference between the corresponding feature sets. This yields an exact mean squared error formula and a sharp bound governed by the energy of the response components that survive cancellation. A complementary lower bound identifies the error that any nonadaptive, equal-weight, reversal-symmetric balanced contrast estimator must incur.

Our contributions are:
\begin{itemize}
    \item \textbf{A balanced interaction estimator with exact nuisance cancellation.} CUBE combines shared model queries with randomized factorial contrasts, preserving pairwise estimates under arbitrary additive and odd-order perturbations.
    \item \textbf{An exact finite-budget error characterization.} We derive closed-form alias probabilities and an exact mean squared error formula, together with an attained worst-case constant determined by the surviving response energy.
    \item \textbf{A matching optimality comparison.} We establish a minimax lower bound for nonadaptive, equal-weight, reversal-symmetric balanced contrast estimators. CUBE attains the same leading constant when its query budget grows while remaining small relative to the full factorial space.
\end{itemize}

\section{Related Work}
\label{sec:related_work}

\subsection{Interaction Definitions and Explanation Targets}
\label{subsec:interaction_definitions}

Feature explanations depend on the model behavior being explained, the treatment of absent features, and the allocation rule~\citep{lundberg2017unified, covert2021explaining, sundararajan2020many}. Conditional and causal formulations incorporate feature dependence and structural knowledge~\citep{aas2021explaining, janzing2020feature, heskes2020causal, frye2020asymmetric}, while global importance can target explained variance or predictive performance~\citep{owen2014sobol, covert2020understanding}. Interaction definitions further differ in coalition weighting and the allocation of higher-order contributions~\citep{grabisch1999axiomatic, sundararajan2020shapley, tsai2023faith, baniecki2026explaining}. Derivative-based methods define contributions along input paths~\citep{sundararajan2017axiomatic, janizek2021explaining}, and functional approaches identify interaction components under a specified distribution~\citep{friedman2008predictive, lengerich2020purifying}, with connections to generalized additive models and cooperative game theory~\citep{bordt2023shapley, fumagalli2024unifying}.

CUBE uses uniform averaging over a fixed two-state probe space, following the connection between Boolean Fourier coefficients and Banzhaf interactions~\citep{grabisch2017note}. The reported effects, $\Delta_i=2\theta_{\{i\}}$ and $\Delta_{ij}=4\theta_{\{i,j\}}$, represent average changes and differences-in-differences for the selected feature states and background.

\subsection{Sampling-Based Interaction Estimation}
\label{subsec:sampling_based_estimation}

Sampling-based attribution approximates exhaustive coalition evaluation using subsets or permutations~\citep{castro2009polynomial, strumbelj2010efficient, vstrumbelj2014explaining}. Error bounds characterize sampling uncertainty~\citep{maleki2013bounding}, while stratification improves allocation across coalition sizes~\citep{castro2017improving, zhang2023efficient}. Multilinear representations express attributions through expectations indexed by feature inclusion probabilities~\citep{okhrati2021multilinear}. Shared measurements also support efficient data valuation and simultaneous estimation of multiple players or probabilistic values~\citep{jia2019towards, kolpaczki2024approximating, wang2023data, li2024one}.

Structured sampling exploits relationships among evaluations through kernel-based point selection, quadrature weights, and orthogonal coupling~\citep{chen2012super, huszar2012optimally, choromanski2019unifying}. For Shapley estimation, structured permutations and paired coalitions improve query efficiency~\citep{mitchell2022sampling, covert2021improving}. OddSHAP further analyzes paired sampling through the odd--even decomposition of set functions, showing that Shapley values depend only on the odd component and using regression on this subspace for estimation~\citep{fumagalli2026odd}. SHAP-IQ and SVARM-IQ extend shared sampling to interaction estimation through general interaction representations and stratification, respectively~\citep{fumagalli2023shap, kolpaczki2024svarm}. The shapiq framework provides implementations and benchmarks with exact interaction targets~\citep{muschalik2024shapiq}.

\subsection{Regression and Sparse Interaction Recovery}
\label{subsec:regression_and_sparse_recovery}

Regression-based explanations fit interpretable representations to model evaluations~\citep{ribeiro2016should}. Regression formulations support Shapley and Banzhaf value estimation, pairwise Shapley interactions, and approximation guarantees based on leverage-score sampling~\citep{liu2024kernel, fumagalli2024kernelshap, musco2024provably}. Residual Monte Carlo correction connects surrogate fitting with attribution estimation~\citep{witter2026regression}. ProxySHAP extends proxy-based estimation with residual correction to Shapley and Banzhaf interactions, combining exact interaction computation for tree ensembles with correction of the proxy's approximation error~\citep{thies2026proxy}. Sparse interaction models use regularization and hierarchy constraints~\citep{bien2013lasso, lim2015learning, haris2016convex, radchenko2010variable}, while tree-based methods identify predictive interaction sets and establish recovery guarantees under specified response models~\citep{shah2014random, basu2018iterative, behr2022provable}.

Spectral methods exploit sparse representations to reduce reconstruction costs through compressed sensing and Walsh--Hadamard algorithms~\citep{candes2006robust, li2015spright}. Related approaches recover sparse set functions and Möbius interactions~\citep{wendler2021learning, kang2024learning}. Compact Fourier representations support amortized attribution and scalable interaction explanations, while tree surrogates enable interaction extraction from queried responses~\citep{gorji2026shap, kang2025spex, butler2026proxyspex}. These methods use joint estimation to separate effects through a fitted representation.

\subsection{Factorial Designs and Randomized Quadrature}
\label{subsec:factorial_designs_and_quadrature}

Fractional factorial designs and orthogonal arrays provide balanced evaluations under limited experimental budgets~\citep{finney1943fractional, plackett1946design, rao1947factorial}. Aberration criteria and indicator-function representations characterize aliasing across regular, nonregular, and mixed-level designs~\citep{fries1980minimum, deng1999generalized, deng1999minimum, wu2001generalized, kenny2003indicator}, while foldover augmentation separates selected aliases through level reversals~\citep{li2003optimal}. Randomized integration connects balance to variance reduction: Latin hypercube sampling stratifies individual coordinates~\citep{mckay2000comparison, stein1987large}, and randomized orthogonal arrays extend coverage to multidimensional projections~\citep{owen1992orthogonal, tang1993orthogonal}, with associated variance analyses and limit theorems~\citep{owen1994lattice, loh1996combinatorial}.

Randomized lattice rules and scrambled nets combine structured coverage with unbiased integration~\citep{cranley1976randomization, owen1997monte}. Discrepancy, smoothness, and local antithetic constructions further characterize and improve integration accuracy~\citep{hickernell1998generalized, owen1997scrambled, owen2008local}. Building on factorial balance, foldover, and randomization, CUBE analyzes the finite-cube integral of $h(z)\phi_S(z)$. Its contribution links explicit collision probabilities to exact interaction-specific MSEs and an attained worst-case risk constant, with a matching asymptotic minimax comparison over nonadaptive, equal-weight, reversal-symmetric balanced contrast estimators.

\section{Method}
\label{sec:method}

For the analysis, we extend the contrast estimator in equation~\ref{eq:estimator} to every $S\subseteq[d]$ using the same empirical average.

\subsection{Explanation Target}
\label{subsec:explanation_target}

Let $f:\mathcal{X}\to\mathbb{R}$ be a fixed predictive model. Assign each selected feature $j\in[d]$ two states $x_j^{-}$ and $x_j^{+}$, with a fixed background for the remaining features. These choices define a probe map $T:\{-1,+1\}^d\to\mathcal{X}$ through $[T(z)]_j=x_j^{-}$ when $z_j=-1$ and $[T(z)]_j=x_j^{+}$ when $z_j=+1$. Write $h(z)=f(T(z))$ and define
\begin{equation}
    \phi_S(z)=\prod_{j\in S}z_j,\qquad
    \theta_S=2^{-d}\sum_{z\in\{-1,+1\}^d}h(z)\phi_S(z),
    \label{eq:target}
\end{equation}
with $\phi_{\emptyset}=1$. CUBE reports $\Delta_i=2\theta_{\{i\}}$ and $\Delta_{ij}=4\theta_{\{i,j\}}$, representing average changes from the designated minus state to the plus state and pairwise differences-in-differences under uniform averaging over the other feature states.

\subsection{Randomized Balanced Queries}
\label{subsec:randomized_balanced_queries}

The construction uses
\begin{equation}
    M=2^r,\qquad B=2M,\qquad d<M,
    \label{eq:budget}
\end{equation}
where $r$ is a positive integer and $B$ counts indexed query rows, including repetitions. Sample ordered distinct labels $u_1,\ldots,u_d$ uniformly from $\mathbb{F}_2^r\setminus\{0\}$ and, independently, uniform signs $\sigma_1,\ldots,\sigma_d\in\{-1,+1\}$. For every $t\in\mathbb{F}_2^r$ and $\eta\in\{-1,+1\}$, query
\begin{equation}
    z_j(\eta,t)=\eta\sigma_j(-1)^{\langle u_j,t\rangle},
    \qquad y_{\eta,t}=h(z(\eta,t)),
    \label{eq:design}
\end{equation}
with binary inner products evaluated modulo two.

Distinct labels give pairwise orthogonality, and $\eta$ pairs each configuration with its global reversal. The resulting design is balanced on every collection of up to three factors. Repeated probes retain their multiplicities in the estimator; caching their responses can reduce the number of distinct model evaluations below $B$.

The theoretical analysis samples labels from all of $\mathbb{F}_2^r$, which induces the same distribution of query multisets. A common translation $u_j\mapsto u_j+c$ only permutes the reversal-paired rows through $\eta\mapsto\eta(-1)^{\langle c,t\rangle}$, and every distinct $d$-tuple has exactly $M-d$ translations that avoid zero.

\subsection{Effect Estimation}
\label{subsec:effect_estimation}

For $\mathcal{I}_2=\{S\subseteq[d]:1\leq|S|\leq2\}$, compute
\begin{equation}
    \widehat{\theta}_S=\frac{1}{B}\sum_{t\in\mathbb{F}_2^r}\sum_{\eta\in\{-1,+1\}}y_{\eta,t}\phi_S(z(\eta,t)),
    \qquad \widehat{\Delta}_S=2^{|S|}\widehat{\theta}_S.
    \label{eq:estimator}
\end{equation}
All target effects reuse the same responses. Reversal pairs make main-effect estimates depend on $[h(z)-h(-z)]/2$ and pairwise estimates depend on $[h(z)+h(-z)]/2$. Algorithm~\ref{alg:cube} summarizes the procedure.

\begin{algorithm}[ht]
\caption{CUBE}
\label{alg:cube}
\begin{algorithmic}[1]
\Require Model $f$, fixed probe map $T$, indexed budget $B=2^{r+1}$ with $d<2^r$
\Ensure Contrasts $\{\widehat{\Delta}_S:S\in\mathcal{I}_2\}$
\State $M\gets B/2$; $\mathcal{I}_2\gets\{S\subseteq[d]:1\leq|S|\leq2\}$
\State Sample ordered distinct $u_1,\ldots,u_d\in\mathbb{F}_2^r\setminus\{0\}$ uniformly
\State Independently sample uniform signs $\sigma_1,\ldots,\sigma_d\in\{-1,+1\}$
\State $\widehat{\theta}_S\gets0$ for every $S\in\mathcal{I}_2$
\For{$t\in\mathbb{F}_2^r$}
    \For{$\eta\in\{-1,+1\}$}
        \State $z_j\gets\eta\sigma_j(-1)^{\langle u_j,t\rangle}$ for every $j\in[d]$
        \State $y\gets f(T(z))$ \Comment{Reuse a cached response when available}
        \State $\widehat{\theta}_S\gets\widehat{\theta}_S+y\phi_S(z)/B$ for every $S\in\mathcal{I}_2$
    \EndFor
\EndFor
\State \Return $\{\widehat{\Delta}_S=2^{|S|}\widehat{\theta}_S:S\in\mathcal{I}_2\}$
\end{algorithmic}
\end{algorithm}

\section{Theory}
\label{sec:theory}

We analyze the fixed response $h:\{-1,+1\}^d\to\mathbb{R}$ under the construction in Section~\ref{sec:method}, with $M=2^r$, $B=2M$, and $d<M$. Expectations concern the random labels $U=(u_1,\ldots,u_d)$ and independent uniform signs $\sigma$. We use the exact expansion $h=\sum_{R\subseteq[d]}\theta_R\phi_R$ and write $\sigma_A=\prod_{j\in A}\sigma_j$.

\subsection{Cancellation and Randomized Aliasing}
\label{subsec:cancellation_and_randomized_aliasing}

For fixed labels, define
\begin{equation}
    \mathcal{K}(U)=\left\{A\subseteq[d]:|A|\text{ is even},\ \bigoplus_{j\in A}u_j=0\right\},
    \label{eq:cube_alias_collection}
\end{equation}
where $\oplus$ denotes binary addition.

\begin{theorem}[Cancellation and unbiasedness]
\label{thm:cube_randomized_aliasing}
For every $S\subseteq[d]$,
\begin{equation}
    \widehat{\theta}_S-\theta_S
    =\sum_{A\in\mathcal{K}(U)\setminus\{\emptyset\}}\sigma_A\theta_{S\triangle A},
    \qquad
    \mathbb{E}_{\sigma}[\widehat{\theta}_S\mid U]=\theta_S.
    \label{eq:cube_randomized_alias_identity}
\end{equation}
For any constant $a_0$ and odd function $g$, satisfying $g(-z)=-g(z)$, every realized design satisfies
\begin{equation}
    \widehat{\theta}_{\{i,j\}}(h+a_0+g)=\widehat{\theta}_{\{i,j\}}(h),
    \qquad i\neq j.
    \label{eq:cube_nuisance_invariance}
\end{equation}
\end{theorem}

\begin{proof}
The empirical average of a character factorizes as
\begin{align}
    m_D(A)
    &:=\frac{1}{B}\sum_{\eta,t}\phi_A(z(\eta,t))\\
    &=\sigma_A
    \left(\frac12\sum_{\eta\in\{-1,+1\}}\eta^{|A|}\right)
    \left(\frac1M\sum_{t\in\mathbb{F}_2^r}(-1)^{\langle\bigoplus_{j\in A}u_j,t\rangle}\right)\\
    &=\sigma_A\mathbf{1}\{A\in\mathcal{K}(U)\}.
    \label{eq:cube_design_character_average}
\end{align}
Expanding $h$ and using $\phi_R\phi_S=\phi_{R\triangle S}$ gives
\begin{equation}
    \widehat{\theta}_S=\sum_R\theta_Rm_D(R\triangle S),
\end{equation}
which proves the alias representation. Every nonempty $\sigma_A$ has mean zero, establishing conditional unbiasedness.

Distinct labels imply $m_D(\{i,j\})=0$, so constants cancel from pairwise contrasts. The contribution of $g$ also vanishes within each reversal pair:
\begin{equation}
    g(z)\phi_{\{i,j\}}(z)+g(-z)\phi_{\{i,j\}}(-z)=0.
\end{equation}
This proves \eqref{eq:cube_nuisance_invariance}.
\end{proof}

The invariance includes arbitrary main effects and all odd-order factorial components. The remaining aliases determine the randomization error.

\subsection{Collision Probabilities}
\label{subsec:collision_probabilities}

Uniform sampling of distinct labels makes $\Pr_U(A\in\mathcal{K}(U))$ depend only on $s=|A|$. Denote this probability by $p_s(B)$.

\begin{proposition}[Exact alias probabilities]
\label{prop:cube_collision_probabilities}
For $0\leq s\leq d<M$,
\begin{equation}
    p_s(B)=
    \begin{cases}
        1, & s=0,\\
        0, & s\text{ is odd},\\[2pt]
        \displaystyle\frac{1}{M}\left[1+(M-1)(-1)^{s/2}\frac{\binom{M/2}{s/2}}{\binom{M}{s}}\right], & s\geq2\text{ is even}.
    \end{cases}
    \label{eq:cube_exact_collision_probability}
\end{equation}
In particular, $p_2(B)=0$ and, when $d\geq4$, $p_4(B)=1/(M-3)$.
\end{proposition}

\begin{proof}
The empty and odd cases follow from \eqref{eq:cube_alias_collection}. For even $s$, character orthogonality yields
\begin{equation}
    p_s(B)=\frac1M\sum_{v\in\mathbb{F}_2^r}\mathbb{E}_U\left[\prod_{j\in A}(-1)^{\langle v,u_j\rangle}\right].
\end{equation}
For each nonzero $v$, exactly half the labels have character value $+1$ and half have value $-1$. Uniform sampling of distinct labels therefore gives
\begin{align}
    \mathbb{E}_U\left[\prod_{j\in A}(-1)^{\langle v,u_j\rangle}\right]
    &=\frac{[x^s](1+x)^{M/2}(1-x)^{M/2}}{\binom Ms}\\
    &=(-1)^{s/2}\frac{\binom{M/2}{s/2}}{\binom Ms},
\end{align}
where $[x^s]$ extracts a coefficient. Including the contribution of $v=0$ proves the formula. Substituting $s=2$ and $s=4$ gives the stated special cases.
\end{proof}

\subsection{Exact Error and Sharp Bound}
\label{subsec:exact_error_and_sharp_bound}

Define the surviving character differences and their response energy by
\begin{equation}
    \mathcal{T}_d=\{A\subseteq[d]:|A|\geq4,\ |A|\text{ is even}\},
    \qquad
    V_S(h)=\sum_{A\in\mathcal{T}_d}\theta_{S\triangle A}^2.
    \label{eq:cube_surviving_energy}
\end{equation}

\begin{theorem}[Exact error and attained worst-case constant]
\label{thm:cube_sharp_error_bound}
For every $S\subseteq[d]$,
\begin{equation}
    \mathbb{E}[(\widehat{\theta}_S-\theta_S)^2]
    =\sum_{R\neq S}p_{|S\triangle R|}(B)\theta_R^2.
    \label{eq:cube_exact_mse}
\end{equation}
For $M\geq4$,
\begin{equation}
    \mathbb{E}[(\widehat{\theta}_S-\theta_S)^2]\leq\frac{2}{B-6}V_S(h).
    \label{eq:cube_uniform_mse_bound}
\end{equation}
If $d\geq4$, the normalized worst-case risk satisfies
\begin{equation}
    \mathcal{R}_{\mathrm{CUBE}}(d,B;S)
    :=\sup_{h:\,V_S(h)>0}\frac{\mathbb{E}[(\widehat{\theta}_S-\theta_S)^2]}{V_S(h)}
    =\frac{2}{B-6}.
    \label{eq:cube_exact_worst_case_risk}
\end{equation}
\end{theorem}

\begin{proof}
Sign products satisfy $\mathbb{E}_{\sigma}[\sigma_A\sigma_{A'}]=\mathbf{1}\{A=A'\}$. Squaring the alias representation and averaging gives
\begin{align}
    \mathbb{E}[(\widehat{\theta}_S-\theta_S)^2]
    &=\mathbb{E}_U\sum_{A\in\mathcal{K}(U)\setminus\{\emptyset\}}\theta_{S\triangle A}^2\\
    &=\sum_{A\in\mathcal{T}_d}p_{|A|}(B)\theta_{S\triangle A}^2.
    \label{eq:cube_error_by_difference}
\end{align}
Reindexing by $R=S\triangle A$ proves \eqref{eq:cube_exact_mse}.

For even $s$, write $a_s=\binom{M/2}{s/2}/\binom Ms$. When $s\equiv2\pmod4$, the probability formula gives $p_s(B)\leq1/M$. For $M\geq8$ and $s\equiv0\pmod4$ with $4\leq s<M$, we have $s\leq M-4$ and
\begin{equation}
    a_{M-s}=a_s,
    \qquad
    \frac{a_{s+2}}{a_s}=\frac{s+1}{M-s-1}.
\end{equation}
Symmetry and the ratio imply $a_s\leq a_4$ on this interval. Consequently,
\begin{equation}
    p_s(B)\leq\frac{1+(M-1)a_4}{M}
    =\frac{1}{M-3}.
\end{equation}
For $M=4$, the condition $d<M$ leaves only $p_2=0$ among positive even indices. Applying these bounds to \eqref{eq:cube_error_by_difference} proves \eqref{eq:cube_uniform_mse_bound}.

For sharpness, choose a four-element set $A_0$ and take $h=\phi_{S\triangle A_0}$. Then $V_S(h)=1$ and
\begin{equation}
    \mathbb{E}[(\widehat{\theta}_S-\theta_S)^2]=p_4(B)=\frac{1}{M-3}=\frac{2}{B-6}.
\end{equation}
\end{proof}

The quantity $V_S(h)$ describes the model's surviving response energy. For the reported contrast $\widehat{\Delta}_S=2^{|S|}\widehat{\theta}_S$, the mean squared error is multiplied by $4^{|S|}$.

\subsection{Minimax Comparison}
\label{subsec:minimax_comparison}

Let $\mathfrak{A}_B$ contain nonadaptive, equal-weight contrast procedures that draw an indexed base design $D_0=(x_1,\ldots,x_B)$ independently of $h$. Each design is invariant under global reversal, including multiplicities, and satisfies $B^{-1}\sum_b x_{b,i}x_{b,j}=0$ for $i\neq j$. After independent uniform level permutations $\sigma$, the procedure returns
\begin{equation}
    \widehat{\theta}^{\mathcal{A}}_S
    =\frac1B\sum_{b=1}^B h(\sigma\odot x_b)\phi_S(\sigma\odot x_b),
    \label{eq:cube_comparison_estimator}
\end{equation}
where $\odot$ denotes coordinatewise multiplication. CUBE belongs to $\mathfrak{A}_B$. Define
\begin{equation}
    \mathfrak{R}_B(d;S)
    =\inf_{\mathcal{A}\in\mathfrak{A}_B}\sup_{h:\,V_S(h)>0}
    \frac{\mathbb{E}[(\widehat{\theta}^{\mathcal{A}}_S-\theta_S)^2]}{V_S(h)}.
    \label{eq:cube_minimax_risk}
\end{equation}

\begin{theorem}[Minimax lower bound]
\label{thm:cube_minimax_lower_bound}
For $d\geq4$ and $B\leq2^d$, whenever $\mathfrak{A}_B$ is nonempty,
\begin{equation}
    \mathfrak{R}_B(d;S)\geq\frac{2^d/B-1}{2^{d-1}-1-\binom d2}.
    \label{eq:cube_minimax_lower_bound}
\end{equation}
\end{theorem}

\begin{proof}
For a base design, define $w_{D_0}(A)=B^{-1}\sum_b\phi_A(x_b)$. Reversal and pairwise balance eliminate all nonconstant moments outside $\mathcal{T}_d$. The same sign-orthogonality argument as above gives
\begin{equation}
    \mathbb{E}[(\widehat{\theta}^{\mathcal{A}}_S-\theta_S)^2]
    =\sum_{A\in\mathcal{T}_d}\theta_{S\triangle A}^2\,\mathbb{E}_{D_0}[w_{D_0}(A)^2].
\end{equation}
The normalized error is a weighted average of expected squared moments, and a single-character response attains the largest one. Thus
\begin{equation}
    \sup_{h:\,V_S(h)>0}
    \frac{\mathbb{E}[(\widehat{\theta}^{\mathcal{A}}_S-\theta_S)^2]}{V_S(h)}
    =\max_{A\in\mathcal{T}_d}\mathbb{E}_{D_0}[w_{D_0}(A)^2].
    \label{eq:cube_general_worst_case}
\end{equation}

Let $N=2^d$ and $\mu(z)=c_z/B$ be the empirical mass of the base design. Parseval's identity gives
\begin{equation}
    \sum_{A\subseteq[d]}w_{D_0}(A)^2
    =N\sum_z\mu(z)^2
    =\frac{N}{B^2}\sum_z c_z^2
    \geq\frac{N}{B},
\end{equation}
using integer counts $c_z$ with $\sum_z c_z=B$. Since the constant moment equals one,
\begin{equation}
    \sum_{A\in\mathcal{T}_d}w_{D_0}(A)^2\geq\frac{N}{B}-1.
\end{equation}
There are $|\mathcal{T}_d|=2^{d-1}-1-\binom d2$ surviving differences. Taking expectations and bounding the maximum by the average yields
\begin{equation}
    \max_{A\in\mathcal{T}_d}\mathbb{E}_{D_0}[w_{D_0}(A)^2]
    \geq\frac{N/B-1}{2^{d-1}-1-\binom d2}.
\end{equation}
Combining this with \eqref{eq:cube_general_worst_case} and taking the infimum over $\mathfrak{A}_B$ proves the result.
\end{proof}

\begin{corollary}[Asymptotic optimality]
\label{cor:cube_near_optimality}
For $d\geq4$ and admissible budgets satisfying $B<2^d$,
\begin{equation}
    \frac{2^d/B-1}{2^{d-1}-1-\binom d2}
    \leq\mathfrak{R}_B(d;S)
    \leq\frac{2}{B-6}.
    \label{eq:cube_minimax_sandwich}
\end{equation}
Along any admissible sequence with $B\to\infty$ and $B/2^d\to0$,
\begin{equation}
    \mathfrak{R}_B(d;S)=\frac{2}{B}(1+o(1)),
    \qquad
    \frac{\mathcal{R}_{\mathrm{CUBE}}(d,B;S)}{\mathfrak{R}_B(d;S)}\to1.
    \label{eq:cube_asymptotic_optimality}
\end{equation}
For $d=5$ and $B=16$, CUBE is exactly minimax within $\mathfrak{A}_{16}$, with normalized risk $1/5$.
\end{corollary}

\begin{proof}
Combine CUBE's attained risk with Theorem~\ref{thm:cube_minimax_lower_bound}. The lower bound can be written as
\begin{equation}
    \frac{2^d/B-1}{2^{d-1}-1-\binom d2}
    =\frac{2}{B}
    \frac{1-B/2^d}{1-(2+2\binom d2)/2^d}.
\end{equation}
Under the stated limits, the last fraction tends to one, as does $B/(B-6)$. The sandwich bound therefore proves \eqref{eq:cube_asymptotic_optimality}. For the finite instance,
\begin{equation}
    \frac{32/16-1}{16-1-\binom52}
    =\frac15
    =\frac{2}{16-6},
\end{equation}
so the lower and upper bounds coincide.
\end{proof}

\section{Experiments}
\label{sec:experiments}

\subsection{Common Experimental Design}
\label{subsec:common_exp_design}

We evaluate pairwise Banzhaf interactions on the coefficient scale $\theta_{\{i,j\}}=\Delta_{ij}/4$. Exact targets and CUBE risks are obtained from the known synthetic coefficients or complete real-model response cubes. Estimators receive only budgeted responses, with repeated inputs charged against the budget and each response shared across estimated pairs. We report mean squared error (MSE) over randomized runs for a focal pair or averaged over all unordered pairs. Experiments~1--2 use unnormalized MSE; Experiments~3--4 divide each nonconstant case's MSE by its response variance before averaging cases. Constant-response cases contribute zero error in Experiment~3 and are excluded from normalized averages in Experiment~4.

The real-model experiments use California Housing, white Wine Quality, and Adult, with respectively $8$, $11$, and $14$ original features, and both XGBoost and multilayer perceptron (MLP) models. Preprocessing is fitted on training data. Each probe switches an original feature between its held-out explained value and its value in a fixed training-set background instance; categorical features switch as complete variables. Responses are regression predictions for California and Wine and positive-class probabilities for Adult. Models, backgrounds, and evaluation cases remain fixed across randomized runs. Further implementation settings are provided in Appendix~\ref{app:reproducibility}.

Experiments~1--3 apply identical CUBE designs to the two responses being compared at budget $B=64$. We construct pointwise $95\%$ confidence intervals using $10{,}000$ bootstrap resamples of design-seed blocks, averaging cases before resampling. Pooled sign-error rates instead use Wilson intervals. These intervals quantify randomization uncertainty conditional on the fixed experimental cases. Small dots in Experiments~1--2 show seed-level estimates; whiskers and shaded bands show confidence intervals. Dashed MSE references in Experiments~1--2 and black diamonds in Experiment~3 indicate exact theoretical predictions.

\subsection{Experiment 1: Matched Outputs and Explanations}
\label{subsec:exp1}

\paragraph{Experimental setup.}
We construct two $12$-feature functions with identical output distributions, main effects, and pairwise coefficients. Both have focal coefficient $\theta_{\{1,2\}}=1$ and a unit fourth-order nuisance coefficient, supported on $\{1,2,3,4\}$ in Function A and $\{1,3,4,5\}$ in Function B. The construction is given in Appendix~\ref{app:matched_outputs_and_risks}. We estimate focal-pair and all-pair MSE using $30$ independent seed blocks, each containing $512$ randomized designs.

\begin{figure}[ht]
    \centering
    \includegraphics[width=\linewidth]{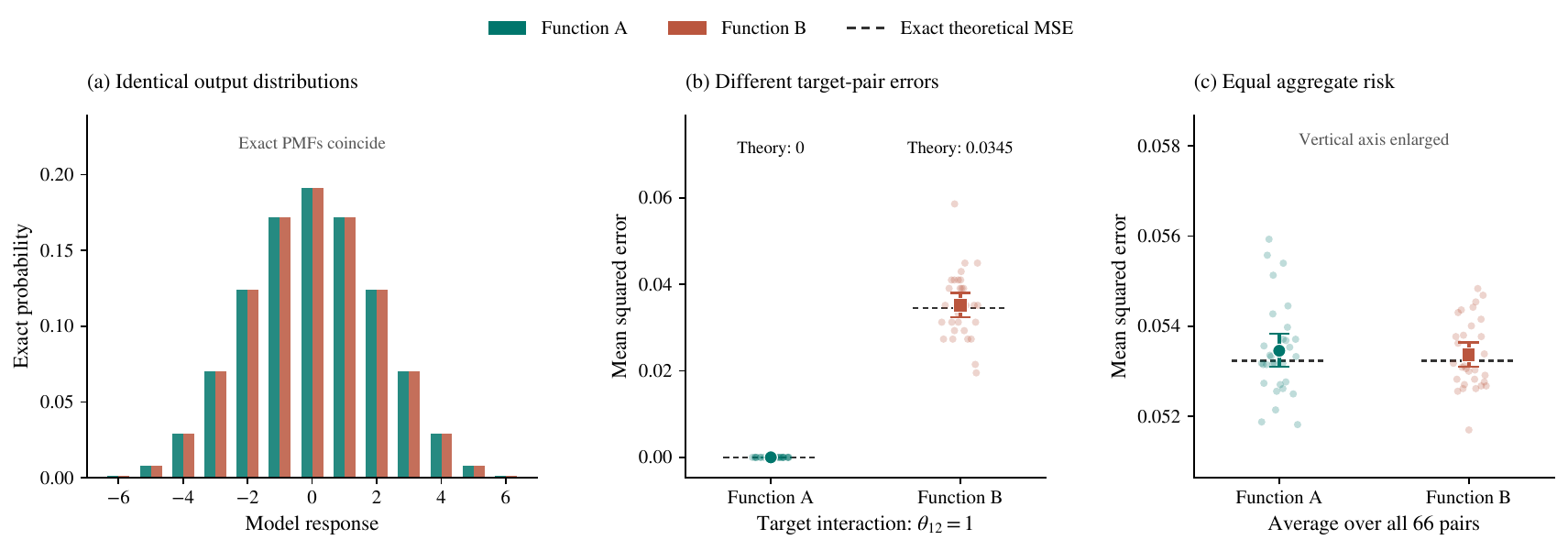}
    \caption{Output distributions, focal-pair errors, and aggregate risk for matched functions.}
    \label{fig:exp1}
\end{figure}

\paragraph{Results and analysis.}
Figure~\ref{fig:exp1} shows different focal risks despite the matched outputs and low-order coefficients. Function A has zero focal error because its nuisance contribution is canceled in every design. Function B has positive focal MSE consistent with the exact value $1/29$. Both all-pair estimates remain close to their common theoretical value, showing that the changed nuisance support redistributes risk across pairs.

\subsection{Experiment 2: Equal MSE and Different Sign Risks}
\label{subsec:exp2}

\paragraph{Experimental setup.}
For $d=12$, we compare
\[
    h_{\mathrm{sp}}(z)=0.8\phi_{\{1,2\}}(z)+\phi_{\{3,4\}}(z),
    \qquad
    h_{\mathrm{de}}(z)=0.8\phi_{\{1,2\}}(z)+\frac{1}{\sqrt{45}}\sum_{3\leq a<b\leq12}\phi_{\{a,b\}}(z).
\]
The nuisance components have equal coefficient energy and equal theoretical focal MSE, $1/29$. Sparse errors take values in $\{0,\pm1\}$, whereas dense errors have magnitude at most $5/\sqrt{45}<0.8$. Consequently, the exact sign-reversal probabilities are $1/58$ and zero, respectively; see Appendix~\ref{app:equal_mse_and_sign_risks}. We use $30$ seed blocks with $1{,}024$ designs per block to compare MSE, absolute-error quantiles, and sign errors.

\begin{figure}[ht]
    \centering
    \includegraphics[width=\linewidth]{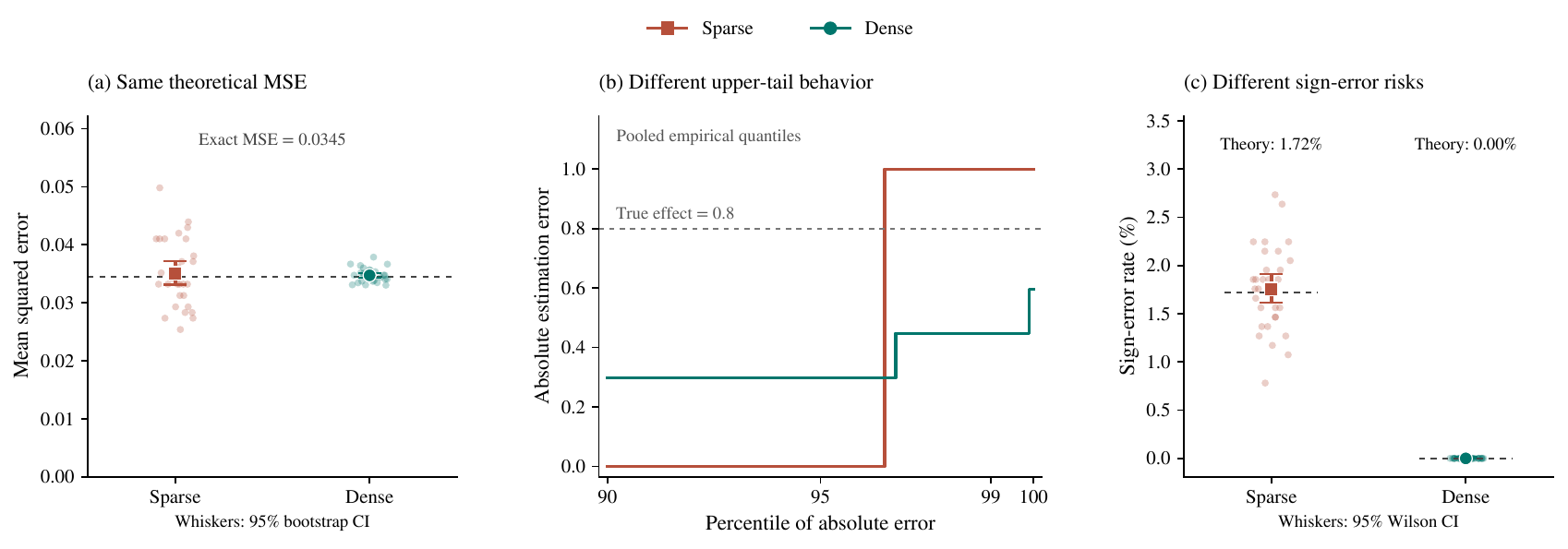}
    \caption{Mean squared error, empirical absolute-error quantiles, and sign errors under sparse and dense nuisance effects.}
    \label{fig:exp2}
\end{figure}

\paragraph{Results and analysis.}
Figure~\ref{fig:exp2} shows similar MSEs but distinct error distributions. Sparse errors are usually zero, with occasional large deviations and greater variation across seed-level MSE estimates. Dense errors occur more frequently at smaller magnitudes. The observed sparse sign-error rate agrees with its theoretical reference, while the dense condition preserves the focal sign in every run. Equal MSE therefore coexists with different tail behavior and sign reliability.

\subsection{Experiment 3: Redistribution of Pairwise Risk}
\label{subsec:exp3}

\paragraph{Experimental setup.}
We evaluate $20$ held-out cases per dataset--model combination. For each complete response cube, we select the pair with the largest absolute exact coefficient and construct a synthetic comparison response by permuting fourth-order coefficients among subsets of features whose two probe states differ. Assigning the largest squared coefficients to supports with the largest focal alias probabilities maximizes focal risk over these permutations. This preserves every other coefficient, the fourth-order coefficient multiset, and response variance. We use $20$ seed blocks with $256$ designs per case and block. The heatmap displays the first selected California/XGBoost case.

\begin{figure}[ht]
    \centering
    \includegraphics[width=\linewidth]{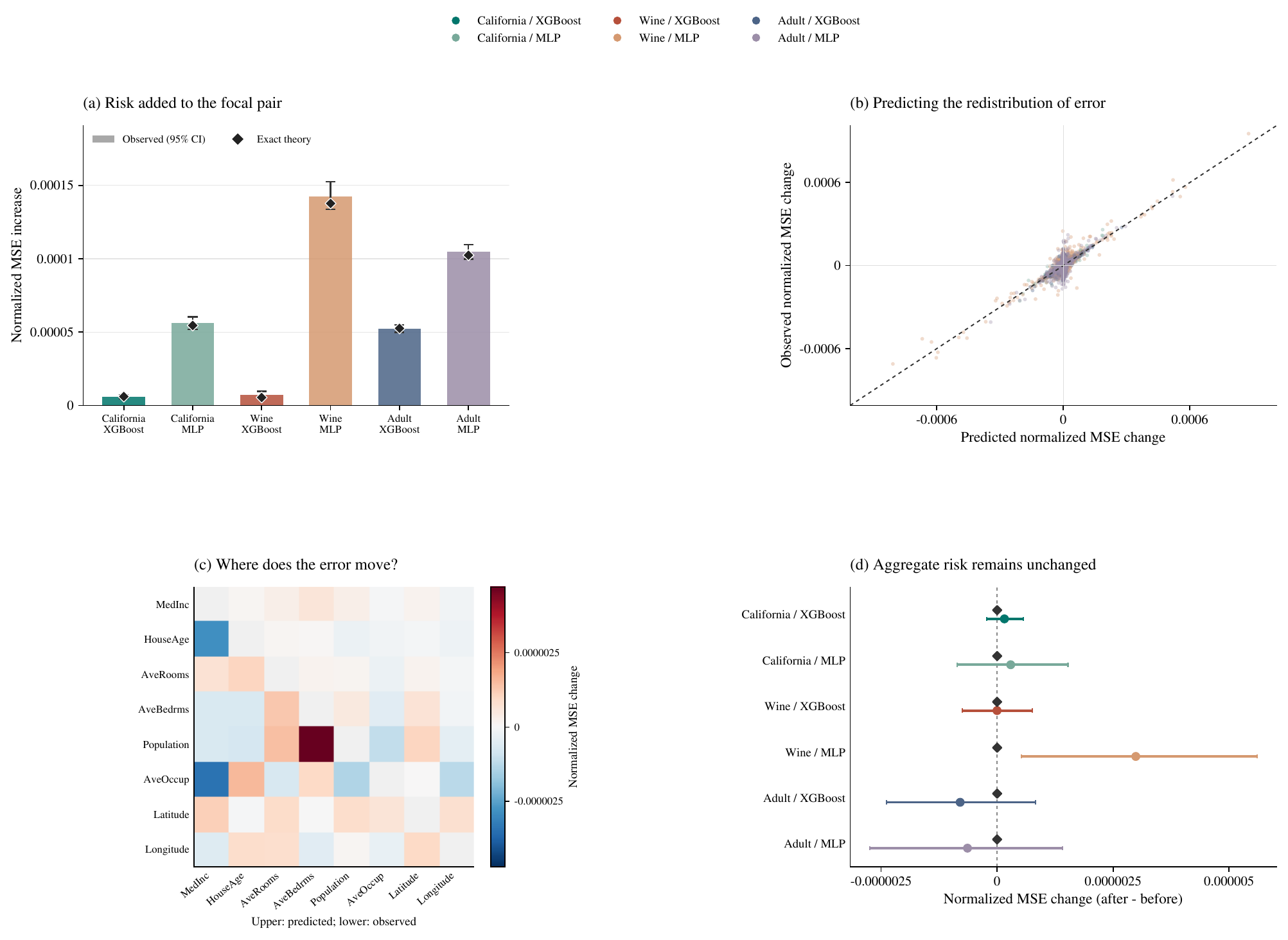}
    \caption{Variance-normalized changes in focal, pairwise, and aggregate MSE after permuting fourth-order coefficients. The heatmap shows predictions above the diagonal and observations below it.}
    \label{fig:exp3}
\end{figure}

\paragraph{Results and analysis.}
Figure~\ref{fig:exp3} shows increased focal risk across all dataset--model combinations, with the largest increases for Wine/MLP and Adult/MLP. Observed pairwise changes largely follow the predicted magnitudes and directions, including both increases and decreases. The all-pair expected change is exactly zero by Appendix~\ref{app:risk_redistribution}. Empirical aggregate changes are small relative to the focal increases; their confidence intervals contain zero except for Wine/MLP, whose small positive estimate has an interval that misses the exact reference.

\subsection{Experiment 4: Equal-Budget Method Comparison}
\label{subsec:exp4}

\paragraph{Experimental setup.}
We compare CUBE with centered paired Monte Carlo, SHAP-IQ, SVARM-IQ, Faith-Banzhaf-2, and ProxySPEX on unmodified real-model responses. Each dataset--model combination uses $10$ held-out cases and $20$ randomized runs per case, method, and budget. Budgets are $\{32,64,128\}$ for California and $\{32,64,128,256,512,1024\}$ for Wine and Adult. Each method receives the same allowed query budget.

Centered paired Monte Carlo draws $m=B/2$ independent uniform configurations $Z_\ell$, queries each configuration and its reversal, and uses the unbiased sample covariance
\begin{equation}
    \widehat{\theta}^{\mathrm{MC}}_S
    =\frac{1}{m-1}\sum_{\ell=1}^{m}(G_\ell-\overline{G})(X_{\ell,S}-\overline{X}_S),
    \qquad
    G_\ell=\frac{h(Z_\ell)+h(-Z_\ell)}{2},
    \quad
    X_{\ell,S}=\phi_S(Z_\ell),
    \label{eq:paired_mc}
\end{equation}
where bars denote sample means and $|S|=2$.

The literature methods use \texttt{shapiq} version~1.7.0. SHAP-IQ and SVARM-IQ target Banzhaf interactions, while Faith-Banzhaf-2 and ProxySPEX use order-two Faith-Banzhaf representations, whose pairwise targets coincide with the same interactions. Their outputs are divided by four to match the coefficient scale. Paired sampling is enabled for SHAP-IQ, SVARM-IQ, and Faith-Banzhaf-2. ProxySPEX selects surrogate depth from $\{2,4,6\}$ by three-fold cross-validation using only its queried responses.

\begin{figure}[ht]
    \centering
    \includegraphics[width=\linewidth]{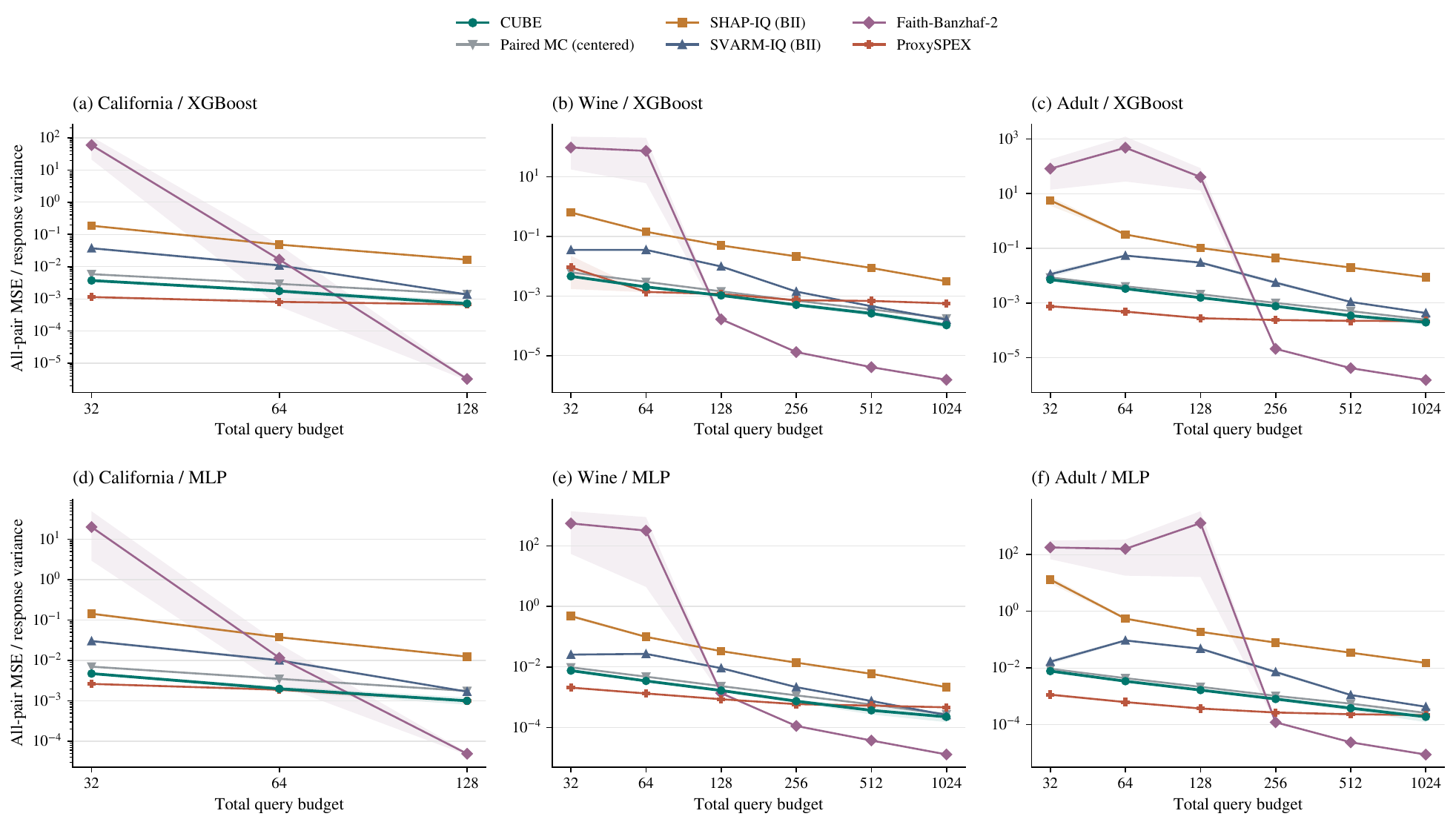}
    \caption{All-pair variance-normalized MSE versus allowed query budget. Columns show California, Wine, and Adult; rows show XGBoost and MLP. Both axes use logarithmic scales.}
    \label{fig:exp4}
\end{figure}

\paragraph{Results and analysis.}
Figure~\ref{fig:exp4} shows decreasing CUBE error with increasing budget, generally below centered paired Monte Carlo, SHAP-IQ, and SVARM-IQ. ProxySPEX achieves lower errors at several small budgets, especially on Adult, but improves more slowly as the budget increases; CUBE overtakes it on Wine and approaches it on Adult. Faith-Banzhaf-2 has large errors at small budgets followed by sharp reductions, achieving the lowest mean error at the largest evaluated budget in every dataset--model combination.

\section{Discussion}
\label{sec:discussion}

\subsection{Target-Relative Estimation Difficulty}
\label{subsec:target_relative_difficulty}

The exact error formula makes estimation difficulty relative to the requested interaction. Main effects and odd-order components contribute no pairwise error, while another pairwise component can contribute when its support is disjoint from the target. Each surviving component contributes its squared coefficient weighted by its alias probability. Consequently, polynomial degree, sparsity, and total response variance alone do not determine the accuracy of a particular interaction estimate.

The matched-function and coefficient-permutation experiments illustrate this dependence. Functions with identical main and pairwise effects can have different focal risks, and rearranging fourth-order coefficients can change individual risks while preserving their all-pair average. Aggregate accuracy therefore gives only a partial account of explanation reliability. Evaluations should also examine how error is distributed across the interactions of interest.

\subsection{Invariance and Unbiasedness as Distinct Guarantees}
\label{subsec:invariance_and_unbiasedness}

Balance controls the influence of nuisance components within each execution. Adding a constant or any response component that is odd under global reversal leaves every pairwise estimate unchanged, regardless of its magnitude. This invariance follows directly from the query arrangement and requires no estimates of the canceled coefficients.

Randomization controls the surviving aliases across executions. The feature labels determine which components collide, and the independent random signs make their error contributions average to zero, even conditional on the labels. The two guarantees describe complementary properties: invariance removes specified sources of error from every estimate, while unbiasedness centers the remaining error distribution. Its spread still depends on the uncanceled response components. The design ablations in Appendix~\ref{app:design_ablation} isolate the effects of reversal, distinct labels, and random signs on cancellation, bias, and MSE.

\subsection{From Estimation Error to Decision Reliability}
\label{subsec:error_and_decision_reliability}

Numerical accuracy and sign reliability depend differently on the target magnitude. Adding $a\phi_S$ shifts both the target coefficient and its contrast estimate by $a$, preserving their difference in every design. MSE is therefore unchanged, while the probability of crossing zero can vary. Directional reliability depends on both the error distribution and the target's distance from zero.

The sparse and dense constructions show that even equal MSE and equal target magnitude can yield different sign risks. Both attain the sharp normalized MSE bound, but one produces rare errors large enough to reverse the sign, whereas the other has a deterministic error bound below the target magnitude. For decisions based on interaction direction, error quantiles and sign-error probabilities provide information that quadratic risk alone cannot supply. Appendices~\ref{app:target_magnitude_sensitivity} and~\ref{app:background_case_sensitivity} examine sign risk across target magnitudes, backgrounds, and evaluation cases.

\subsection{Cancellation and Joint Recovery}
\label{subsec:cancellation_and_joint_recovery}

Balanced contrasts and joint fitting extract information from queried responses in different ways. CUBE imposes cancellation through the design and estimates each effect by a shared-response contrast. Regression fits coefficients jointly, using the sampled relationships among their basis functions. For an exactly quadratic response, noiseless observations with a full-rank quadratic design identify all coefficients. A CUBE design can still alias disjoint pairs, so a joint estimator with an identifying design can achieve smaller errors in this setting.

Experiment~4 shows that Faith-Banzhaf-2 achieves the lowest mean error at the largest tested budgets. Explaining its sharp improvement requires examining both the sampled design rank and the contribution of response components beyond the fitted representation. CUBE's minimax result characterizes worst-case normalized MSE within the specified class of nonadaptive, equal-weight, reversal-symmetric balanced contrast estimators. Its upper and lower bounds match asymptotically as admissible budgets grow while remaining small relative to the full probe space. This comparison leaves room for joint estimators to exploit response structure and achieve lower errors on particular problems.

\subsection{Limitations and Future Work}
\label{subsec:limitations_and_future_work}

CUBE explains a fixed two-state probe space under uniform averaging, so its interpretation depends on the selected feature states and background. Probe configurations may also depart from combinations represented in the data. The construction restricts both query budgets and their relation to feature dimension. Its exact MSE formula involves unknown response coefficients, and the surviving energy $V_S(h)$ generally cannot be recovered from a single budgeted design. The current analysis therefore provides an expected-risk characterization rather than an observable error certificate for an individual execution. Empirical evidence is limited to the evaluated tabular models and cases.

Future work could develop balanced constructions for arbitrary budgets and probe distributions that respect feature constraints. Repeated independent designs offer a route to estimating randomization uncertainty, with further analysis needed for calibrated intervals and simultaneous coverage across pairs. Combining balanced querying with joint estimation could improve recovery while preserving nuisance invariance, motivating guarantees for a broader estimator class. Evaluations across additional model families and explanation settings would clarify how these gains depend on response structure.

\section{Conclusion}
\label{sec:conclusion}

We introduced CUBE to estimate main effects and pairwise interactions from shared balanced queries over a fixed two-state probe space. Each design makes pairwise estimates invariant to constant and odd-order components, including main effects, while randomization ensures unbiasedness. Explicit alias probabilities yield an exact mean squared error formula and a sharp risk bound without truncating response orders. This bound asymptotically matches the minimax constant for nonadaptive, equal-weight, reversal-symmetric balanced contrast estimators as admissible budgets grow while remaining small relative to the full probe space. Experiments show that nuisance structure redistributes interaction-specific risk and that equal mean squared errors can conceal different sign risks. CUBE thus connects exact cancellation, residual aliasing, and attainable accuracy in interaction estimation.

\section*{Broader Impact Statement}

CUBE estimates feature interactions through balanced model queries, combining exact cancellation of selected nuisance effects with an explicit characterization of estimation error. The method supports model inspection over a specified two-state probe space.

More accurate interaction estimates may help practitioners identify unexpected feature dependencies and investigate undesirable model behavior. Reusing evaluations across effects may also reduce the computational cost of such inspections.

Interaction estimates may be misinterpreted as causal relationships or evidence of model fairness. Their meaning depends on the selected feature states and background, which may generate unrealistic inputs. Residual estimation error can also produce misleading signs or rankings, especially for weak interactions.

Practitioners should document the probe construction, assess whether queried inputs are meaningful, and examine stability across repeated designs and relevant backgrounds. Consequential decisions should combine these explanations with domain evidence and broader model evaluation, accounting for uncertainty in individual estimates.

\section*{Reproducibility}

We provide the experimental code in the Supplementary Material as \texttt{CUBE.py}, organized into separately executable Google Colab cells. The code includes CUBE, comparison methods, experimental configurations, evaluation metrics, random seeds, and routines for generating numerical summaries and figures. Execution requirements and implementation settings are described in Appendix~\ref{app:reproducibility}, and runtime package versions are recorded with the outputs.

\bibliography{main}

@inproceedings{lundberg2017unified,
  title={A unified approach to interpreting model predictions},
  author={Lundberg, Scott M and Lee, Su-In},
  booktitle={Advances in Neural Information Processing Systems},
  volume={30},
  year={2017}
}

@article{covert2021explaining,
  title={Explaining by removing: A unified framework for model explanation},
  author={Covert, Ian and Lundberg, Scott and Lee, Su-In},
  journal={Journal of Machine Learning Research},
  volume={22},
  number={209},
  pages={1--90},
  year={2021}
}

@inproceedings{sundararajan2020many,
  title={The many {Shapley} values for model explanation},
  author={Sundararajan, Mukund and Najmi, Amir},
  booktitle={International Conference on Machine Learning},
  pages={9269--9278},
  year={2020},
  organization={PMLR}
}

@article{aas2021explaining,
  title={Explaining individual predictions when features are dependent: More accurate approximations to {Shapley} values},
  author={Aas, Kjersti and Jullum, Martin and L{\o}land, Anders},
  journal={Artificial Intelligence},
  volume={298},
  pages={103502},
  year={2021},
  publisher={Elsevier}
}

@inproceedings{janzing2020feature,
  title={Feature relevance quantification in explainable {AI}: A causal problem},
  author={Janzing, Dominik and Minorics, Lenon and Bl{\"o}baum, Patrick},
  booktitle={International Conference on Artificial Intelligence and Statistics},
  pages={2907--2916},
  year={2020},
  organization={PMLR}
}

@inproceedings{heskes2020causal,
  title={Causal {Shapley} values: Exploiting causal knowledge to explain individual predictions of complex models},
  author={Heskes, Tom and Sijben, Evi and Bucur, Ioan Gabriel and Claassen, Tom},
  booktitle={Advances in Neural Information Processing Systems},
  volume={33},
  pages={4778--4789},
  year={2020}
}

@inproceedings{frye2020asymmetric,
  title={Asymmetric {Shapley} values: incorporating causal knowledge into model-agnostic explainability},
  author={Frye, Christopher and Rowat, Colin and Feige, Ilya},
  booktitle={Advances in Neural Information Processing Systems},
  volume={33},
  pages={1229--1239},
  year={2020}
}

@article{owen2014sobol,
  title={{Sobol'} indices and {Shapley} value},
  author={Owen, Art B},
  journal={SIAM/ASA Journal on Uncertainty Quantification},
  volume={2},
  number={1},
  pages={245--251},
  year={2014},
  publisher={SIAM}
}

@inproceedings{covert2020understanding,
  title={Understanding global feature contributions with additive importance measures},
  author={Covert, Ian and Lundberg, Scott M and Lee, Su-In},
  booktitle={Advances in Neural Information Processing Systems},
  volume={33},
  pages={17212--17223},
  year={2020}
}

@article{grabisch1999axiomatic,
  title={An axiomatic approach to the concept of interaction among players in cooperative games},
  author={Grabisch, Michel and Roubens, Marc},
  journal={International Journal of Game Theory},
  volume={28},
  number={4},
  pages={547--565},
  year={1999},
  publisher={Springer}
}

@inproceedings{sundararajan2020shapley,
  title={The {Shapley} taylor interaction index},
  author={Sundararajan, Mukund and Dhamdhere, Kedar and Agarwal, Ashish},
  booktitle={International Conference on Machine Learning},
  pages={9259--9268},
  year={2020},
  organization={PMLR}
}

@article{tsai2023faith,
  title={{Faith-Shap}: The faithful {Shapley} interaction index},
  author={Tsai, Che-Ping and Yeh, Chih-Kuan and Ravikumar, Pradeep},
  journal={Journal of Machine Learning Research},
  volume={24},
  number={94},
  pages={1--42},
  year={2023}
}

@inproceedings{baniecki2026explaining,
  title={Explaining similarity in vision-language encoders with weighted {Banzhaf} interactions},
  author={Baniecki, Hubert and Muschalik, Maximilian and Fumagalli, Fabian and Hammer, Barbara and H{\"u}llermeier, Eyke and Biecek, Przemyslaw},
  booktitle={Advances in Neural Information Processing Systems},
  volume={38},
  pages={54529--54566},
  year={2026}
}

@inproceedings{sundararajan2017axiomatic,
  title={Axiomatic attribution for deep networks},
  author={Sundararajan, Mukund and Taly, Ankur and Yan, Qiqi},
  booktitle={International Conference on Machine Learning},
  pages={3319--3328},
  year={2017},
  organization={PMLR}
}

@article{janizek2021explaining,
  title={Explaining explanations: Axiomatic feature interactions for deep networks},
  author={Janizek, Joseph D and Sturmfels, Pascal and Lee, Su-In},
  journal={Journal of Machine Learning Research},
  volume={22},
  number={104},
  pages={1--54},
  year={2021}
}

@article{friedman2008predictive,
  title={Predictive learning via rule ensembles},
  author={Friedman, Jerome H and Popescu, Bogdan E},
  year={2008},
  journal={The Annals of Applied Statistics},
  volume={2},
  number={3},
  pages={916--954},
  doi={10.1214/07-AOAS148},
  url={https://arxiv.org/abs/0811.1679},
}

@inproceedings{lengerich2020purifying,
  title={Purifying interaction effects with the functional {ANOVA}: An efficient algorithm for recovering identifiable additive models},
  author={Lengerich, Benjamin and Tan, Sarah and Chang, Chun-Hao and Hooker, Giles and Caruana, Rich},
  booktitle={International Conference on Artificial Intelligence and Statistics},
  pages={2402--2412},
  year={2020},
  organization={PMLR},
  volume={108},
  series={Proceedings of Machine Learning Research},
  url={https://proceedings.mlr.press/v108/lengerich20a.html},
}

@inproceedings{bordt2023shapley,
  title={From {Shapley} values to generalized additive models and back},
  author={Bordt, Sebastian and von Luxburg, Ulrike},
  booktitle={International Conference on Artificial Intelligence and Statistics},
  pages={709--745},
  year={2023},
  organization={PMLR}
}

@article{fumagalli2024unifying,
  title={Unifying feature-based explanations with functional {ANOVA} and cooperative game theory},
  author={Fumagalli, Fabian and Muschalik, Maximilian and H{\"u}llermeier, Eyke and Hammer, Barbara and Herbinger, Julia},
  journal={arXiv preprint arXiv:2412.17152},
  year={2024}
}

@article{grabisch2017note,
  title={A note on the {Sobol'} indices and interactive criteria},
  author={Grabisch, Michel and Labreuche, Christophe},
  journal={Fuzzy Sets and Systems},
  volume={315},
  pages={99--108},
  year={2017},
  publisher={Elsevier}
}

@article{castro2009polynomial,
  title={Polynomial calculation of the {Shapley} value based on sampling},
  author={Castro, Javier and G{\'o}mez, Daniel and Tejada, Juan},
  journal={Computers \& Operations Research},
  volume={36},
  number={5},
  pages={1726--1730},
  year={2009},
  publisher={Elsevier}
}

@article{strumbelj2010efficient,
  title={An efficient explanation of individual classifications using game theory},
  author={{\v{S}}trumbelj, Erik and Kononenko, Igor},
  journal={The Journal of Machine Learning Research},
  volume={11},
  pages={1--18},
  year={2010},
  publisher={JMLR. org}
}

@article{vstrumbelj2014explaining,
  title={Explaining prediction models and individual predictions with feature contributions},
  author={{\v{S}}trumbelj, Erik and Kononenko, Igor},
  journal={Knowledge and Information Systems},
  volume={41},
  number={3},
  pages={647--665},
  year={2014},
  publisher={Springer}
}

@article{maleki2013bounding,
  title={Bounding the estimation error of sampling-based {Shapley} value approximation},
  author={Maleki, Sasan and Tran-Thanh, Long and Hines, Greg and Rahwan, Talal and Rogers, Alex},
  journal={arXiv preprint arXiv:1306.4265},
  year={2013}
}

@article{castro2017improving,
  title={Improving polynomial estimation of the {Shapley} value by stratified random sampling with optimum allocation},
  author={Castro, Javier and G{\'o}mez, Daniel and Molina, Elisenda and Tejada, Juan},
  journal={Computers \& Operations Research},
  volume={82},
  pages={180--188},
  year={2017},
  publisher={Elsevier}
}

@article{zhang2023efficient,
  title={Efficient sampling approaches to {Shapley} value approximation},
  author={Zhang, Jiayao and Sun, Qiheng and Liu, Jinfei and Xiong, Li and Pei, Jian and Ren, Kui},
  journal={Proceedings of the ACM on Management of Data},
  volume={1},
  number={1},
  pages={1--24},
  year={2023},
  publisher={ACM New York, NY, USA}
}

@inproceedings{okhrati2021multilinear,
  title={A multilinear sampling algorithm to estimate {Shapley} values},
  author={Okhrati, Ramin and Lipani, Aldo},
  booktitle={2020 25th International Conference on Pattern Recognition (ICPR)},
  pages={7992--7999},
  year={2021},
  organization={IEEE}
}

@inproceedings{jia2019towards,
  title={Towards efficient data valuation based on the {Shapley} value},
  author={Jia, Ruoxi and Dao, David and Wang, Boxin and Hubis, Frances Ann and Hynes, Nick and G{\"u}rel, Nezihe Merve and Li, Bo and Zhang, Ce and Song, Dawn and Spanos, Costas J},
  booktitle={Proceedings of the 22nd International Conference on Artificial Intelligence and Statistics},
  pages={1167--1176},
  year={2019},
  organization={PMLR}
}

@article{chen2012super,
  title={Super-samples from kernel herding},
  author={Chen, Yutian and Welling, Max and Smola, Alex},
  journal={arXiv preprint arXiv:1203.3472},
  year={2012}
}

@article{huszar2012optimally,
  title={Optimally-weighted herding is {Bayesian} quadrature},
  author={Husz{\'a}r, Ferenc and Duvenaud, David},
  journal={arXiv preprint arXiv:1204.1664},
  year={2012}
}

@inproceedings{choromanski2019unifying,
  title={Unifying orthogonal {Monte Carlo} methods},
  author={Choromanski, Krzysztof and Rowland, Mark and Chen, Wenyu and Weller, Adrian},
  booktitle={International Conference on Machine Learning},
  pages={1203--1212},
  year={2019},
  organization={PMLR}
}

@article{mitchell2022sampling,
  title={Sampling permutations for {Shapley} value estimation},
  author={Mitchell, Rory and Cooper, Joshua and Frank, Eibe and Holmes, Geoffrey},
  journal={Journal of Machine Learning Research},
  volume={23},
  number={43},
  pages={1--46},
  year={2022}
}

@inproceedings{covert2021improving,
  title={Improving {KernelSHAP}: Practical {Shapley} value estimation using linear regression},
  author={Covert, Ian and Lee, Su-In},
  booktitle={International Conference on Artificial Intelligence and Statistics},
  pages={3457--3465},
  year={2021},
  organization={PMLR}
}

@inproceedings{kolpaczki2024approximating,
  title={Approximating the {Shapley} value without marginal contributions},
  author={Kolpaczki, Patrick and Bengs, Viktor and Muschalik, Maximilian and H{\"u}llermeier, Eyke},
  booktitle={Proceedings of the AAAI Conference on Artificial Intelligence},
  volume={38},
  pages={13246--13255},
  year={2024}
}

@inproceedings{wang2023data,
  title={Data {Banzhaf}: A robust data valuation framework for machine learning},
  author={Wang, Jiachen T and Jia, Ruoxi},
  booktitle={International Conference on Artificial Intelligence and Statistics},
  pages={6388--6421},
  year={2023},
  organization={PMLR}
}

@inproceedings{li2024one,
  title={One sample fits all: Approximating all probabilistic values simultaneously and efficiently},
  author={Li, Weida and Yu, Yaoliang},
  booktitle={Advances in Neural Information Processing Systems},
  volume={37},
  pages={58309--58340},
  year={2024}
}

@inproceedings{fumagalli2023shap,
  title={{SHAP-IQ}: Unified approximation of any-order {Shapley} interactions},
  author={Fumagalli, Fabian and Muschalik, Maximilian and Kolpaczki, Patrick and H{\"u}llermeier, Eyke and Hammer, Barbara},
  booktitle={Advances in Neural Information Processing Systems},
  volume={36},
  pages={11515--11551},
  year={2023}
}

@inproceedings{kolpaczki2024svarm,
  title={{SVARM-IQ}: Efficient approximation of any-order {Shapley} interactions through stratification},
  author={Kolpaczki, Patrick and Muschalik, Maximilian and Fumagalli, Fabian and Hammer, Barbara and H{\"u}llermeier, Eyke},
  year={2024},
  booktitle={Proceedings of the 27th International Conference on Artificial Intelligence and Statistics},
  volume={238},
  series={Proceedings of Machine Learning Research},
  pages={3520--3528},
  publisher={PMLR},
  url={https://proceedings.mlr.press/v238/kolpaczki24a.html},
}

@inproceedings{muschalik2024shapiq,
  title={{shapiq}: {Shapley} interactions for machine learning},
  author={Muschalik, Maximilian and Baniecki, Hubert and Fumagalli, Fabian and Kolpaczki, Patrick and Hammer, Barbara and H{\"u}llermeier, Eyke},
  booktitle={Advances in Neural Information Processing Systems},
  volume={37},
  pages={130324--130357},
  year={2024}
}

@inproceedings{ribeiro2016should,
  title={``Why should {I} trust you?'' Explaining the predictions of any classifier},
  author={Ribeiro, Marco Tulio and Singh, Sameer and Guestrin, Carlos},
  booktitle={Proceedings of the 22nd ACM SIGKDD International Conference on Knowledge Discovery and Data Mining},
  pages={1135--1144},
  year={2016}
}

@article{liu2024kernel,
  title={Kernel {Banzhaf}: A Fast and Robust Estimator for {Banzhaf} Values},
  author={Liu, Yurong and Witter, R Teal and Korn, Flip and Alrashed, Tarfah and Paparas, Dimitris and Musco, Christopher and Freire, Juliana},
  journal={arXiv preprint arXiv:2410.08336},
  year={2024}
}

@article{fumagalli2024kernelshap,
  title={{KernelSHAP-IQ}: Weighted least-square optimization for {Shapley} interactions},
  author={Fumagalli, Fabian and Muschalik, Maximilian and Kolpaczki, Patrick and H{\"u}llermeier, Eyke and Hammer, Barbara},
  journal={arXiv preprint arXiv:2405.10852},
  year={2024}
}

@article{musco2024provably,
  title={Provably accurate {Shapley} value estimation via leverage score sampling},
  author={Musco, Christopher and Witter, R Teal},
  journal={arXiv preprint arXiv:2410.01917},
  year={2024}
}

@inproceedings{witter2026regression,
  title={Regression-adjusted {Monte Carlo} estimators for {Shapley} values and probabilistic values},
  author={Witter, R Teal and Liu, Yurong and Musco, Christopher},
  booktitle={Advances in Neural Information Processing Systems},
  volume={38},
  pages={21209--21240},
  year={2026}
}

@article{bien2013lasso,
  title={A lasso for hierarchical interactions},
  author={Bien, Jacob and Taylor, Jonathan and Tibshirani, Robert},
  journal={The Annals of Statistics},
  volume={41},
  number={3},
  pages={1111--1141},
  year={2013},
  doi={10.1214/13-AOS1096},
  url={https://arxiv.org/abs/1205.5050},
}

@article{lim2015learning,
  title={Learning interactions via hierarchical group-lasso regularization},
  author={Lim, Michael and Hastie, Trevor},
  journal={Journal of Computational and Graphical Statistics},
  volume={24},
  number={3},
  pages={627--654},
  year={2015},
  publisher={Taylor \& Francis}
}

@article{haris2016convex,
  title={Convex modeling of interactions with strong heredity},
  author={Haris, Asad and Witten, Daniela and Simon, Noah},
  journal={Journal of Computational and Graphical Statistics},
  volume={25},
  number={4},
  pages={981--1004},
  year={2016},
  publisher={Taylor \& Francis}
}

@article{radchenko2010variable,
  title={Variable selection using adaptive nonlinear interaction structures in high dimensions},
  author={Radchenko, Peter and James, Gareth M},
  journal={Journal of the American Statistical Association},
  volume={105},
  number={492},
  pages={1541--1553},
  year={2010},
  publisher={Taylor \& Francis}
}

@article{shah2014random,
  title={Random intersection trees},
  author={Shah, Rajen Dinesh and Meinshausen, Nicolai},
  journal={The Journal of Machine Learning Research},
  volume={15},
  number={1},
  pages={629--654},
  year={2014},
  publisher={JMLR. org}
}

@article{basu2018iterative,
  title={Iterative random forests to discover predictive and stable high-order interactions},
  author={Basu, Sumanta and Kumbier, Karl and Brown, James B and Yu, Bin},
  journal={Proceedings of the National Academy of Sciences},
  volume={115},
  number={8},
  pages={1943--1948},
  year={2018},
  publisher={National Academy of Sciences}
}

@article{behr2022provable,
  title={Provable {Boolean} interaction recovery from tree ensemble obtained via random forests},
  author={Behr, Merle and Wang, Yu and Li, Xiao and Yu, Bin},
  journal={Proceedings of the National Academy of Sciences},
  volume={119},
  number={22},
  pages={e2118636119},
  year={2022},
  publisher={National Academy of Sciences}
}

@article{candes2006robust,
  title={Robust uncertainty principles: Exact signal reconstruction from highly incomplete frequency information},
  author={Cand{\`e}s, Emmanuel J and Romberg, Justin and Tao, Terence},
  journal={IEEE Transactions on Information Theory},
  volume={52},
  number={2},
  pages={489--509},
  year={2006},
  publisher={IEEE}
}

@article{li2015spright,
  title={{SPRIGHT}: A fast and robust framework for sparse {Walsh-Hadamard} transform},
  author={Li, Xiao and Bradley, Joseph K and Pawar, Sameer and Ramchandran, Kannan},
  journal={arXiv preprint arXiv:1508.06336},
  year={2015}
}

@inproceedings{wendler2021learning,
  title={Learning set functions that are sparse in non-orthogonal {Fourier} bases},
  author={Wendler, Chris and Amrollahi, Andisheh and Seifert, Bastian and Krause, Andreas and P{\"u}schel, Markus},
  booktitle={Proceedings of the AAAI Conference on Artificial Intelligence},
  volume={35},
  number={12},
  pages={10283--10292},
  year={2021}
}

@inproceedings{kang2024learning,
  title={Learning to understand: Identifying interactions via the M{\"o}bius transform},
  author={Kang, Justin S and Erginbas, Yigit E and Butler, Landon and Pedarsani, Ramtin and Ramchandran, Kannan},
  booktitle={Advances in Neural Information Processing Systems},
  volume={37},
  pages={46160--46202},
  year={2024}
}

@inproceedings{gorji2026shap,
  title={{SHAP} values via sparse {Fourier} representation},
  author={Gorji, Ali and Amrollahi, Andisheh and Krause, Andreas},
  booktitle={Advances in Neural Information Processing Systems},
  volume={38},
  pages={18827--18860},
  year={2026}
}

@article{kang2025spex,
  title={{SPEX}: Scaling feature interaction explanations for {LLMs}},
  author={Kang, Justin Singh and Butler, Landon and Agarwal, Abhineet and Erginbas, Yigit Efe and Pedarsani, Ramtin and Ramchandran, Kannan and Yu, Bin},
  journal={arXiv preprint arXiv:2502.13870},
  year={2025}
}

@inproceedings{butler2026proxyspex,
  title={{ProxySPEX}: Inference-efficient interpretability via sparse feature interactions in {LLMs}},
  author={Butler, Landon and Agarwal, Abhineet and Kang, Justin and Erginbas, Yigit Efe and Yu, Bin and Ramchandran, Kannan},
  booktitle={Advances in Neural Information Processing Systems},
  volume={38},
  pages={72306--72340},
  year={2026}
}

@article{finney1943fractional,
  title={The fractional replication of factorial arrangements},
  author={Finney, D. J.},
  journal={Annals of Eugenics},
  volume={12},
  number={1},
  pages={291--301},
  year={1943},
}

@article{plackett1946design,
  title={The design of optimum multifactorial experiments},
  author={Plackett, Robin L and Burman, J Peter},
  journal={Biometrika},
  volume={33},
  number={4},
  pages={305--325},
  year={1946},
}

@article{rao1947factorial,
  title={Factorial experiments derivable from combinatorial arrangements of arrays},
  author={Rao, C Radhakrishna},
  journal={Supplement to the Journal of the Royal Statistical Society},
  volume={9},
  number={1},
  pages={128--139},
  year={1947},
}

@article{fries1980minimum,
  title={Minimum aberration {$2^{k-p}$} designs},
  author={Fries, Arthur and Hunter, William G},
  journal={Technometrics},
  volume={22},
  number={4},
  pages={601--608},
  year={1980},
  publisher={Taylor \& Francis}
}

@article{deng1999generalized,
  title={Generalized resolution and minimum aberration criteria for {Plackett-Burman} and other nonregular factorial designs},
  author={Deng, Lih-Yuan and Tang, Boxin},
  journal={Statistica Sinica},
  pages={1071--1082},
  year={1999},
  volume={9},
  number={4},
}

@article{deng1999minimum,
  title={Minimum {$G_2$}-aberration for nonregular fractional factorial designs},
  author={Tang, Boxin and Deng, Lih-Yuan},
  journal={The Annals of Statistics},
  volume={27},
  number={6},
  pages={1914--1926},
  year={1999},
  publisher={Institute of Mathematical Statistics},
  doi={10.1214/aos/1017939244},
}

@article{wu2001generalized,
  title={Generalized minimum aberration for asymmetrical fractional factorial designs},
  author={Xu, Hongquan and Wu, C. F. J.},
  journal={The Annals of Statistics},
  volume={29},
  number={4},
  pages={1066--1077},
  year={2001},
  publisher={Institute of Mathematical Statistics},
  doi={10.1214/aos/1013699993},
}

@article{kenny2003indicator,
  title={Indicator function and its application in two-level factorial designs},
  author={Ye, Kenny Q.},
  journal={The Annals of Statistics},
  volume={31},
  number={3},
  pages={984--994},
  year={2003},
  publisher={Institute of Mathematical Statistics},
  doi={10.1214/aos/1056562470},
}

@article{li2003optimal,
  title={Optimal foldover plans for two-level fractional factorial designs},
  author={Li, William and Lin, Dennis K J},
  journal={Technometrics},
  volume={45},
  number={2},
  pages={142--149},
  year={2003},
  publisher={Taylor \& Francis}
}

@article{mckay2000comparison,
  title={A comparison of three methods for selecting values of input variables in the analysis of output from a computer code},
  author={McKay, Michael D and Beckman, Richard J and Conover, William J},
  journal={Technometrics},
  volume={42},
  number={1},
  pages={55--61},
  year={2000},
  publisher={Taylor \& Francis}
}

@article{stein1987large,
  title={Large sample properties of simulations using {Latin} hypercube sampling},
  author={Stein, Michael},
  journal={Technometrics},
  volume={29},
  number={2},
  pages={143--151},
  year={1987},
  publisher={Taylor \& Francis}
}

@article{owen1992orthogonal,
  title={Orthogonal arrays for computer experiments, integration and visualization},
  author={Owen, Art B},
  journal={Statistica Sinica},
  pages={439--452},
  year={1992},
  volume={2},
  number={2},
}

@article{tang1993orthogonal,
  title={Orthogonal array-based {Latin} hypercubes},
  author={Tang, Boxin},
  journal={Journal of the American Statistical Association},
  volume={88},
  number={424},
  pages={1392--1397},
  year={1993},
  publisher={Taylor \& Francis}
}

@article{owen1994lattice,
  title={Lattice sampling revisited: {Monte Carlo} variance of means over randomized orthogonal arrays},
  author={Owen, Art},
  journal={The Annals of Statistics},
  pages={930--945},
  year={1994},
  volume={22},
  number={2},
  doi={10.1214/aos/1176325504},
}

@article{loh1996combinatorial,
  title={A combinatorial central limit theorem for randomized orthogonal array sampling designs},
  author={Loh, Wei-Liem},
  journal={The Annals of Statistics},
  volume={24},
  number={3},
  pages={1209--1224},
  year={1996},
  publisher={Institute of Mathematical Statistics}
}

@article{cranley1976randomization,
  title={Randomization of number theoretic methods for multiple integration},
  author={Cranley, Roy and Patterson, Thomas NL},
  journal={SIAM Journal on Numerical Analysis},
  volume={13},
  number={6},
  pages={904--914},
  year={1976},
  publisher={SIAM}
}

@article{owen1997monte,
  title={{Monte Carlo} variance of scrambled net quadrature},
  author={Owen, Art B},
  journal={SIAM Journal on Numerical Analysis},
  volume={34},
  number={5},
  pages={1884--1910},
  year={1997},
  publisher={SIAM}
}

@article{hickernell1998generalized,
  title={A generalized discrepancy and quadrature error bound},
  author={Hickernell, Fred},
  journal={Mathematics of Computation},
  volume={67},
  number={221},
  pages={299--322},
  year={1998}
}

@article{owen1997scrambled,
  title={Scrambled net variance for integrals of smooth functions},
  author={Owen, Art B},
  journal={The Annals of Statistics},
  volume={25},
  number={4},
  pages={1541--1562},
  year={1997},
  publisher={Institute of Mathematical Statistics}
}

@article{owen2008local,
  title={Local antithetic sampling with scrambled nets},
  author={Owen, Art B},
  year={2008},
  journal={The Annals of Statistics},
  volume={36},
  number={5},
  pages={2319--2343},
  doi={10.1214/07-AOS548},
  url={https://arxiv.org/abs/0811.0528},
}

@article{fumagalli2026odd,
  title   = {An Odd Estimator for {Shapley} Values},
  author  = {Fumagalli, Fabian and Butler, Landon and Kang, Justin Singh
             and Ramchandran, Kannan and Witter, R. Teal},
  journal = {arXiv preprint arXiv:2602.01399},
  year    = {2026},
  url     = {https://arxiv.org/abs/2602.01399}
}

@article{thies2026proxy,
  title   = {Proxy-Based Approximation of {Shapley} and {Banzhaf} Interactions},
  author  = {Thies, Santo M. A. R. and Baniecki, Hubert and Witter, R. Teal
             and H{\"u}llermeier, Eyke and Muschalik, Maximilian
             and Fumagalli, Fabian},
  journal = {arXiv preprint arXiv:2605.22738},
  year    = {2026},
  url     = {https://arxiv.org/abs/2605.22738}
}
\bibliographystyle{tmlr}

\appendix

\section{Proofs}
\label{app:proofs}

\subsection{Bounding Alias Collision Probabilities}
\label{app:alias_probability_bound}

We detail the uniform bound on the alias probabilities used in Theorem~4.3. The argument controls the binomial coefficient ratio in the exact probability formula, including the boundary case $M=4$.

\begin{lemma}
Let $M=2^r\ge 4$, $B=2M$, and $d<M$. For every $1\le s\le d$,
\[
p_s(B)\le \frac{1}{M-3}=\frac{2}{B-6}.
\]
If $d\ge 4$, equality holds at $s=4$.
\end{lemma}

\begin{proof}
Odd indices and $s=2$ satisfy $p_s(B)=0$. For positive even $s<M$, define
\[
a_s=\frac{\binom{M/2}{s/2}}{\binom{M}{s}},
\qquad
p_s(B)=\frac{1+(M-1)(-1)^{s/2}a_s}{M}.
\]
When $s\equiv 2\pmod 4$, positivity of $a_s$ gives
\[
p_s(B)\le \frac{1}{M}<\frac{1}{M-3}.
\]

Suppose $M\ge 8$ and $s\equiv 0\pmod 4$. Since $M$ is divisible by four and $s<M$, we have $4\le s\le M-4$. Binomial symmetry and cancellation of factorials yield
\[
a_{M-s}=a_s,
\qquad
\frac{a_{k+2}}{a_k}
=
\frac{M-k}{k+2}
\frac{(k+1)(k+2)}{(M-k)(M-k-1)}
=
\frac{k+1}{M-k-1}
\]
for even $0\le k\le M-2$. Set $t=\min\{s,M-s\}$. Then $t$ is even, $4\le t\le M/2$, and $a_s=a_t$. For each even $k$ with $4\le k<t$, we have $k\le M/2-2$, so the displayed ratio is at most one. Consequently,
\[
a_s=a_t\le a_4
=
\frac{3}{(M-1)(M-3)}.
\]
Substitution into the probability formula gives
\[
p_s(B)
\le
\frac{1+(M-1)a_4}{M}
=
\frac{1}{M-3}.
\]

For $M=4$, the condition $d<M$ leaves only odd indices and $s=2$, whose probabilities vanish. Finally, if $d\ge 4$, the index $s=4$ is admissible and direct substitution gives $p_4(B)=1/(M-3)$, establishing equality.
\end{proof}

\subsection{Matched Output Distributions and Distinct Interaction Risks}
\label{app:matched_outputs_and_risks}

For $d\ge5$, define
\[
g(z)=\frac12\sum_{j=5}^{d}z_j,
\qquad
h_A(z)=g(z)+\phi_{\{1,2\}}(z)+\phi_{\{1,2,3,4\}}(z),
\qquad
h_B(z)=g(z)+\phi_{\{1,2\}}(z)+\phi_{\{1,3,4,5\}}(z).
\]
Experiment~1 uses $d=12$ and $B=64$.

\begin{proposition}
Under uniform cube inputs, $h_A$ and $h_B$ have identical output distributions, main effects, and pairwise coefficients. For the focal pair $S=\{1,2\}$ and every admissible CUBE budget,
\[
\widehat\theta_S(h_A)=\theta_S(h_A)=1
\]
in every realized design, whereas
\[
\mathbb E\!\left[
\bigl(\widehat\theta_S(h_B)-\theta_S(h_B)\bigr)^2
\right]
=p_4(B)=\frac{2}{B-6}.
\]
Their all-pair expected MSEs are equal.
\end{proposition}

\begin{proof}
Let $Z$ be uniform on $\{-1,+1\}^d$, and write
\[
W=(Z_5,\ldots,Z_d),
\qquad
T=Z_1Z_2,
\qquad
E=Z_3Z_4.
\]
Conditional on $(Z_1,W)$, $T$ and $E$ are independent uniform signs. The target and nuisance terms are $(T,TE)$ for $h_A$ and $(T,Z_1Z_5E)$ for $h_B$. Both maps from $(T,E)$ are bijections of $\{-1,+1\}^2$, so each resulting pair consists of independent uniform signs conditional on $(Z_1,W)$. Consequently, for $C\in\{A,B\}$,
\[
\mathcal L(h_C(Z)\mid W)
=
\frac14\delta_{g(W)-2}
+\frac12\delta_{g(W)}
+\frac14\delta_{g(W)+2},
\]
where $g(W)=\frac12\sum_{j=5}^{d}Z_j$ and $\delta_v$ denotes a point mass at $v$. Averaging over $W$ establishes equality of the output distributions.

Character orthogonality gives, for both responses,
\[
\theta_{\{j\}}=\frac12\mathbf 1\{j\ge5\},
\qquad
\theta_{\{i,j\}}=\mathbf 1\{\{i,j\}=\{1,2\}\}.
\]
Thus their main effects and pairwise coefficients coincide. In the focal contrast, reversal symmetry cancels $g$, and the target term contributes exactly one. The nuisance supports satisfy
\[
S\triangle\{1,2,3,4\}=\{3,4\},
\qquad
S\triangle\{1,3,4,5\}=\{2,3,4,5\}.
\]
Pairwise balance cancels the first contribution in every design. The second has squared coefficient one and symmetric-difference size four, so the exact MSE formula yields $p_4(B)=2/(B-6)$, equal to $1/29$ at $B=64$.

To establish aggregate equality, let
\[
\overline{\mathcal R}_2(h)
=
\binom d2^{-1}
\sum_{\substack{Q\subseteq[d]\\|Q|=2}}
\mathbb E\!\left[
\bigl(\widehat\theta_Q(h)-\theta_Q(h)\bigr)^2
\right].
\]
The shared main effects contribute zero to every pairwise risk. The unit coefficient on $S$ contributes $p_4(B)$ to each of the $\binom{d-2}{2}$ pairs disjoint from $S$ and zero to all other pairs. For either fourth-order support $R$, a pair $Q$ has symmetric-difference size
\[
|Q\triangle R|=6-2|Q\cap R|.
\]
There are $\binom{d-4}{2}$ pairs with no overlap, $4(d-4)$ with one shared feature, and $\binom42$ contained in $R$. Their respective contributions are $p_6(B)$, $p_4(B)$, and $p_2(B)=0$. Terms with zero combinatorial multiplicity are omitted; in particular, the $p_6(B)$ term is absent when $d=5$. Hence, for both $C=A$ and $C=B$,
\[
\overline{\mathcal R}_2(h_C)
=
\frac{
\left[\binom{d-2}{2}+4(d-4)\right]p_4(B)
+\binom{d-4}{2}p_6(B)
}{
\binom d2
}.
\]
This expression depends on the nuisance support only through its cardinality, proving the claimed equality.
\end{proof}

\subsection{Equal Mean Squared Errors and Different Sign Risks}
\label{app:equal_mse_and_sign_risks}

Let $d\ge 5$, $n=d-2$, and $K=\binom{n}{2}$. For a positive target coefficient $\mu$ and nuisance magnitude $\lambda>0$, Experiment~2 compares
\[
h_{\mathrm{sp}}(z)
=
\mu\phi_{\{1,2\}}(z)+\lambda\phi_{\{3,4\}}(z),
\qquad
h_{\mathrm{de}}(z)
=
\mu\phi_{\{1,2\}}(z)
+
\frac{\lambda}{\sqrt K}
\sum_{3\le a<b\le d}\phi_{\{a,b\}}(z).
\]
Both nuisance components have coefficient energy $\lambda^2$. Write $S=\{1,2\}$ and $e_C=\widehat\theta_S(h_C)-\mu$ for $C\in\{\mathrm{sp},\mathrm{de}\}$.

\begin{proposition}
For any admissible CUBE budget,
\[
\mathbb E[e_{\mathrm{sp}}^2]
=
\mathbb E[e_{\mathrm{de}}^2]
=
\lambda^2p_4(B)
=
\frac{2\lambda^2}{B-6}.
\]
The sparse error satisfies
\[
\Pr(e_{\mathrm{sp}}=0)=1-p_4(B),
\qquad
\Pr(e_{\mathrm{sp}}=\lambda)
=
\Pr(e_{\mathrm{sp}}=-\lambda)
=
\frac{p_4(B)}{2},
\]
whereas every realized design satisfies
\[
|e_{\mathrm{de}}|
\le
L_n,
\qquad
L_n=\frac{\lambda\lfloor n/2\rfloor}{\sqrt K}.
\]
Consequently, if $L_n<\mu<\lambda$, then
\[
\Pr\!\left(\widehat\theta_S(h_{\mathrm{sp}})<0\right)
=
\frac{p_4(B)}{2},
\qquad
\Pr\!\left(\widehat\theta_S(h_{\mathrm{de}})<0\right)=0.
\]
\end{proposition}

\begin{proof}
Every nuisance pair is disjoint from $S$, so its symmetric difference with $S$ has size four. The exact MSE formula therefore gives
\[
\mathbb E[e_{\mathrm{sp}}^2]=\lambda^2p_4(B),
\qquad
\mathbb E[e_{\mathrm{de}}^2]
=
K\frac{\lambda^2}{K}p_4(B).
\]
In both cases, $V_S(h_C)=\lambda^2$, so the normalized risks attain the sharp bound.

For the sparse response, the alias representation reduces to
\[
e_{\mathrm{sp}}
=
\lambda\sigma_1\sigma_2\sigma_3\sigma_4
\mathbf 1\{u_1\oplus u_2=u_3\oplus u_4\}.
\]
The collision event has probability $p_4(B)$ and depends only on the labels. The sign product is an independent uniform sign, establishing the stated error distribution.

For the dense response, define the colliding nuisance pairs by
\[
\mathcal E(U)
=
\bigl\{\{a,b\}:3\le a<b\le d,\;
u_a\oplus u_b=u_1\oplus u_2\bigr\}.
\]
Then
\[
e_{\mathrm{de}}
=
\frac{\lambda}{\sqrt K}\sigma_1\sigma_2
\sum_{\{a,b\}\in\mathcal E(U)}\sigma_a\sigma_b.
\]
For each $a$, the collision condition determines its partner's label uniquely as $u_b=u_a\oplus u_1\oplus u_2$. Distinct labels therefore prevent two colliding pairs from sharing a feature. Thus $\mathcal E(U)$ is a matching on $n$ features, and
\[
|e_{\mathrm{de}}|
\le
\frac{\lambda}{\sqrt K}|\mathcal E(U)|
\le
\frac{\lambda\lfloor n/2\rfloor}{\sqrt K}.
\]
If $L_n<\mu<\lambda$, the dense estimate remains strictly positive, while the sparse estimate is negative exactly when $e_{\mathrm{sp}}=-\lambda$.
\end{proof}

The experiment uses $d=12$, $B=64$, $\lambda=1$, and $\mu=0.8$. Here $L_n=5/\sqrt{45}<0.8<1$, giving a common MSE of $1/29$ and sign-error probabilities $1/58$ and zero, respectively. The separation follows from the error distributions of these fixed constructions despite their equal quadratic risks.

\subsection{Risk Redistribution and Aggregate Error Preservation}
\label{app:risk_redistribution}

Let $\mathcal P_2=\{S\subseteq[d]:|S|=2\}$, and define the pairwise and aggregate risks by
\[
\mathcal R_S(h)
=
\mathbb E\!\left[(\widehat\theta_S-\theta_S)^2\right],
\qquad
\overline{\mathcal R}(h)
=
\frac{1}{\binom d2}\sum_{S\in\mathcal P_2}\mathcal R_S(h).
\]
Write $E_k(h)=\sum_{|R|=k}\theta_R^2$. To exclude each target coefficient from its own error, set $\kappa_0=0$, $\kappa_s=p_s(B)$ for $1\le s\le d$, and $\kappa_s=0$ outside this range.

\begin{proposition}
For any response $h$ and admissible CUBE budget,
\[
\overline{\mathcal R}(h)
=
\sum_{k=0}^{d}w_k E_k(h),
\qquad
w_k
=
\frac{
\binom{k}{2}\kappa_{k-2}
+k(d-k)\kappa_k
+\binom{d-k}{2}\kappa_{k+2}
}{\binom d2}.
\]
Consequently, preserving the coefficient energy at every order preserves the aggregate expected MSE.
\end{proposition}

\begin{proof}
The exact MSE formula and interchange of finite sums give
\[
\overline{\mathcal R}(h)
=
\frac{1}{\binom d2}
\sum_{R\subseteq[d]}\theta_R^2
\sum_{S\in\mathcal P_2}\kappa_{|S\triangle R|}.
\]
Fix $R$ with $|R|=k$. A pair $S$ can intersect $R$ in two, one, or zero features. The corresponding numbers of pairs are $\binom{k}{2}$, $k(d-k)$, and $\binom{d-k}{2}$, and their symmetric differences have sizes $k-2$, $k$, and $k+2$, respectively. Thus the inner sum depends only on $k$. Grouping coefficients by their order proves the identity.
\end{proof}

For Experiment~3, let $\mathcal F$ contain the fourth-order subsets eligible for permutation, restricted to features with distinct probe states. A permutation $\pi$ of $\mathcal F$ defines
\[
\theta'_R=
\begin{cases}
\theta_{\pi(R)}, & R\in\mathcal F,\\
\theta_R, & R\notin\mathcal F,
\end{cases}
\qquad
h'=\sum_{R\subseteq[d]}\theta'_R\phi_R.
\]
Every pairwise coefficient remains unchanged, and the exact change in its estimation risk is
\[
\mathcal R_S(h')-\mathcal R_S(h)
=
\sum_{R\in\mathcal F}
p_{|S\triangle R|}(B)
\bigl(\theta_{\pi(R)}^2-\theta_R^2\bigr).
\]
The permutation preserves each $E_k$, so the proposition yields
\[
\overline{\mathcal R}(h')=\overline{\mathcal R}(h),
\qquad
\sum_{S\in\mathcal P_2}
\bigl(\mathcal R_S(h')-\mathcal R_S(h)\bigr)=0.
\]

The implementation selects a focal pair $S_\star$ with maximal $|\theta_S|$ and assigns squared coefficients in increasing order to increasing weights $p_{|S_\star\triangle R|}(B)$. This assignment maximizes the focal risk over the permitted permutations. Indeed, assigning $a\le b$ to weights $u\le v$ instead of the opposite assignment changes their contribution by
\[
(au+bv)-(bu+av)=(b-a)(v-u)\ge 0.
\]
Successive exchanges establish the maximizing order. Since the original assignment is feasible, the focal risk cannot decrease; any strict increase is offset by decreases across other pairs.

Finally, Parseval's identity gives
\[
\operatorname{Var}(h'(Z))
=
\sum_{R\ne\varnothing}(\theta'_R)^2
=
\sum_{R\ne\varnothing}\theta_R^2
=
\operatorname{Var}(h(Z)).
\]
Aggregate preservation therefore also holds after normalization by the common positive response variance. These equalities concern expected risks over randomized designs; empirical averages over finitely many designs may differ.

\section{Ablation and Sensitivity Analyses}
\label{app:ablation_sensitivity_analyses}

\subsection{Common Experimental Design}
\label{app:common_exp_design}

The synthetic experiments use responses $h(z)=\mu\phi_S(z)+\lambda g(z)$ on the uniform binary cube, where $S=\{1,2\}$ and $g$ has unit coefficient energy and no component on $S$. Unless varied explicitly, $d=12$, $B=64$, and $\mu=\lambda=1$. CUBE uses distinct nonzero labels, independent random signs, and global reversal, with $B=2M$, $M$ a power of two, and $d<M$. Query budgets count indexed rows with multiplicity. Errors are measured on the coefficient scale; conversion to $\Delta_{ij}=4\theta_{ij}$ multiplies MSE by $16$ and preserves sign and tie probabilities.

MSE is unnormalized in the design-ablation and target-magnitude experiments, divided by nuisance energy $\lambda^2$ in the nuisance-structure experiment, and divided by surviving energy $V_S(h)=\lambda^2$ in the dimension-budget experiment. Real-model MSE is normalized by each probe's response variance, with constant probes excluded from normalized averages. Exact references are computed from known synthetic spectra or complete response cubes and are used for evaluation.

Each experiment uses $30$ independent seed blocks. The numbers of designs per block are $512$, $512$, $1024$, $256$, and $32$, respectively, in the order presented below. Related conditions share randomization where applicable; magnitude sweeps reuse query-evaluated errors through exact linearity. Pointwise $95\%$ intervals for MSE, bias, paired differences, and the ablation error probabilities use $10{,}000$ whole-seed bootstrap resamples. Synthetic sign, tie, and exceedance probabilities in the remaining experiments use Wilson intervals over independent designs. Real-model sign rates use the seed-block bootstrap, preserving dependence across pairs and probes. All intervals describe randomization uncertainty conditional on the fixed experimental conditions.

\subsection{Design Ablation and Nuisance Magnitude Sensitivity}
\label{app:design_ablation}

\paragraph{Experimental setup.}
We compare CUBE with variants that omit global reversal, fix all random signs to $+1$, or sample nonzero labels with replacement. The variant without global reversal applies two independent sign vectors to the same label array, retaining $B$ query rows. All variants use the uncentered contrast. We vary $\lambda$ over $\{0,0.1,0.3,1,3,10,30,100\}$ for $g(z)\in\{1,z_3,z_1z_3z_4,z_3z_4\}$. We measure target MSE, bias for the disjoint pair $g(z)=z_3z_4$, and its unconditional error probabilities at $\lambda=1$.

\begin{figure}[ht]
    \centering
    \includegraphics[width=\linewidth]{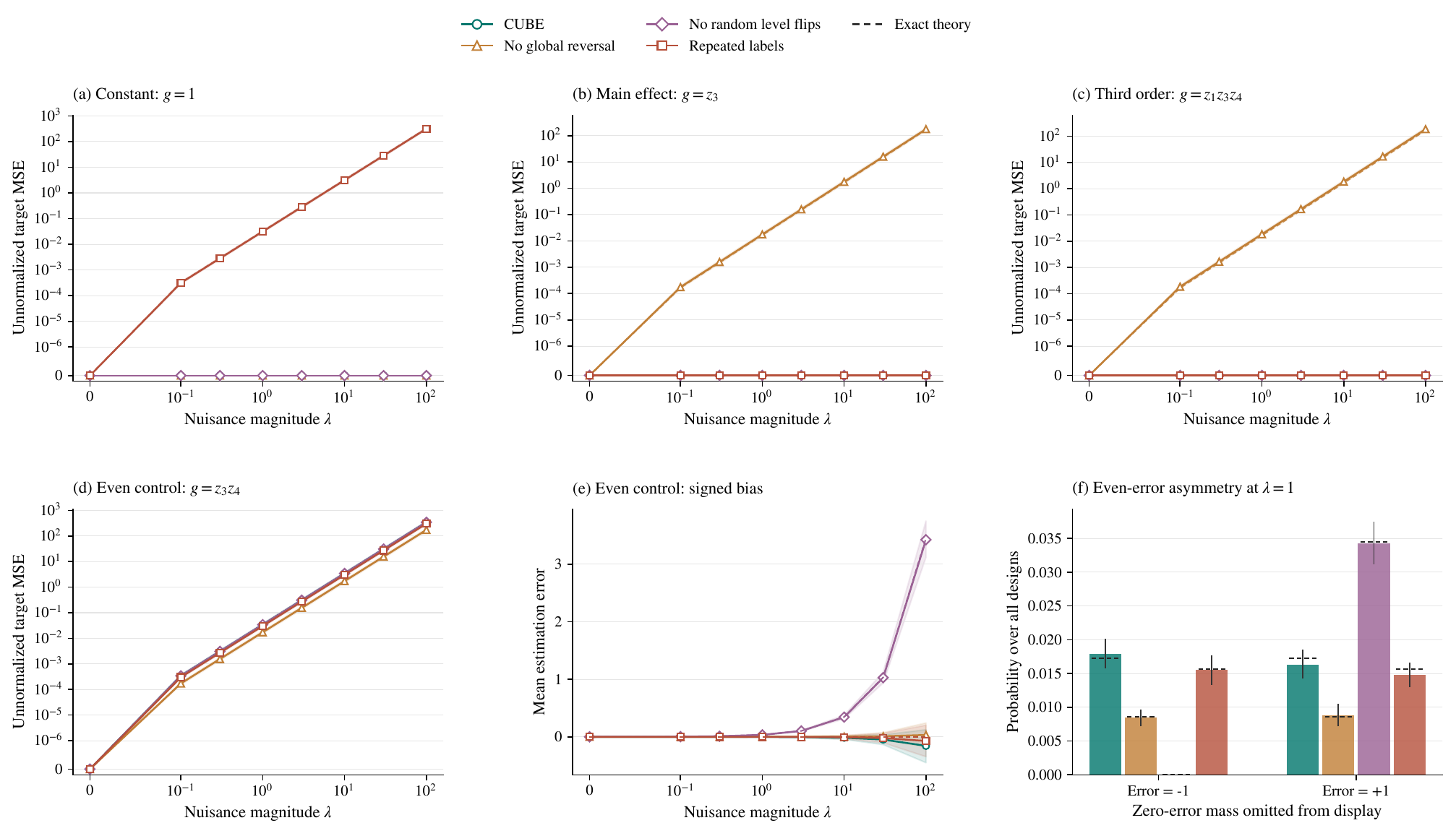}
    \caption{Design ablation: (a--d) target MSE across nuisance magnitudes; (e) signed bias for the disjoint pair; (f) unconditional probabilities of errors $-1$ and $+1$ at $\lambda=1$. Zero-error mass is omitted from panel (f). Dashed marks indicate exact references.}
    \label{fig:app_design_ablation}
\end{figure}

\paragraph{Results and analysis.}
Figure~\ref{fig:app_design_ablation} shows constant leakage under repeated labels and main-effect and third-order leakage without global reversal. These contributions produce MSE proportional to $\lambda^2$, while CUBE cancels them in every design. For the disjoint pair, replacing reversal with an independent sign vector reduces MSE from $\lambda^2p_4(B)$ to $\lambda^2p_4(B)/2$. Fixing the signs instead preserves CUBE's MSE but introduces bias $\lambda p_4(B)$. At $\lambda=1$, its nonzero errors are exclusively positive, whereas CUBE assigns equal probability to errors $-1$ and $+1$. Thus the ablations distinguish cancellation, unbiasedness, and quadratic risk.

\subsection{Sensitivity to Nuisance Structure}
\label{app:nuisance_structure_sensitivity}

\paragraph{Experimental setup.}
We vary nuisance structure at fixed target coefficient and total nuisance energy. The first family uses eight equal-magnitude terms for each feasible combination of order $k\in\{2,4,6,8\}$ and target overlap $r\in\{0,1,2\}$, excluding the target itself. Every term then has symmetric-difference size $k+2-2r$, giving exact MSE $\lambda^2p_{k+2-2r}(B)$. The second family transfers fourth-order energy from overlap $r=2$ to $r=1$, varying the surviving fraction $\eta$ from zero to one. The third distributes unit energy over nested collections of $1$, $2$, $4$, $8$, $16$, or $32$ fourth-order terms with $r=1$. Supports and coefficient signs remain fixed across randomized designs. We evaluate normalized MSE and absolute-error exceedances at $t/\lambda\in\{0.25,0.75,1.25\}$.

\begin{figure}[ht]
    \centering
    \includegraphics[width=\linewidth]{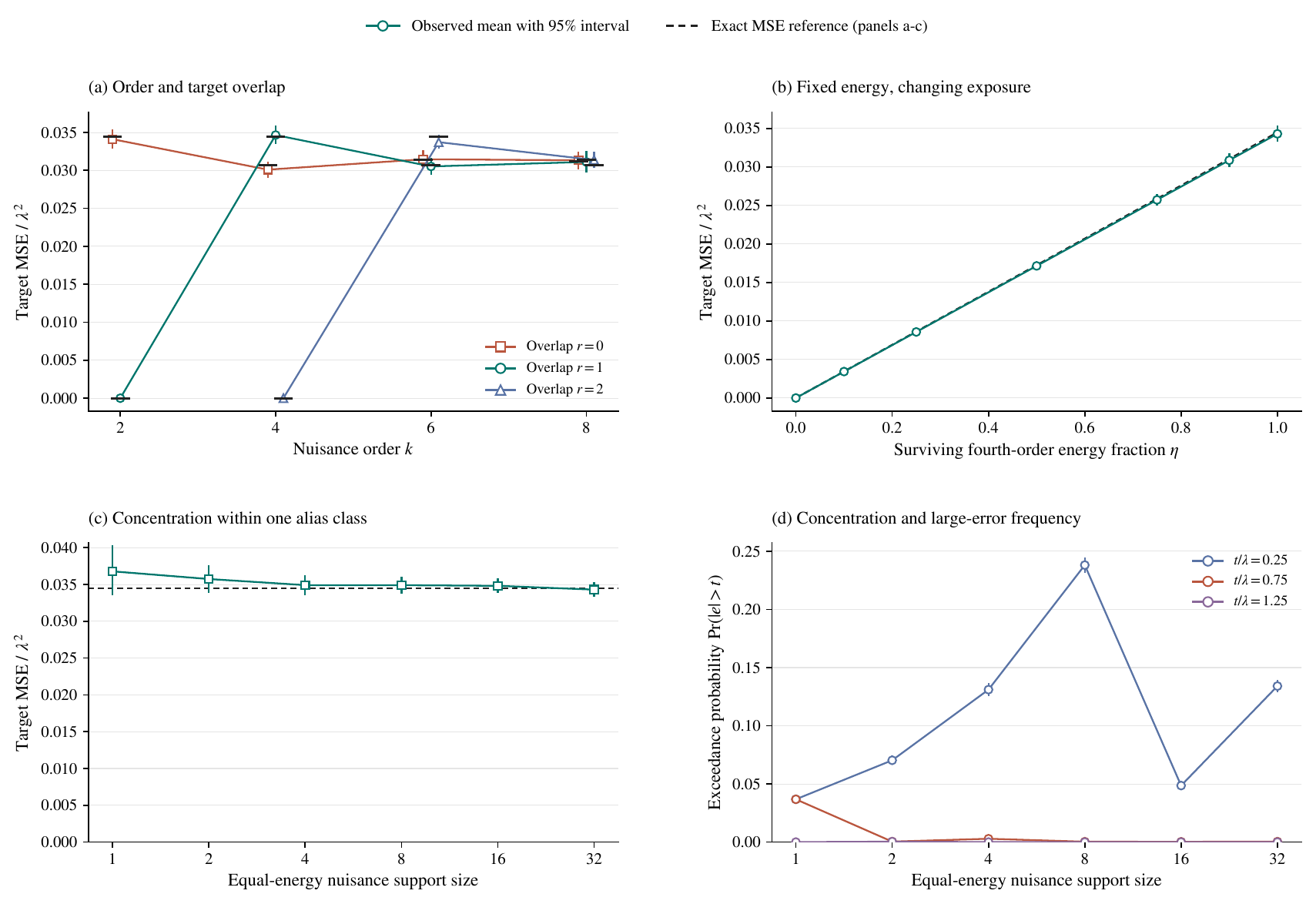}
    \caption{Nuisance structure at fixed energy: normalized MSE versus (a) order and target overlap, (b) surviving energy fraction, and (c) nuisance-term count at equal alias probability (the panel's ``alias class''); (d) absolute-error exceedance probabilities at three thresholds. Exact MSE references appear in (a--c).}
    \label{fig:app_nuisance_structure}
\end{figure}

\paragraph{Results and analysis.}
The feasible order-overlap combinations in Figure~\ref{fig:app_nuisance_structure} show exact cancellation when $k+2-2r=2$ and positive MSE otherwise. Risk does not increase uniformly with nuisance order. Transferring energy to surviving components produces the predicted linear change $\eta p_4(B)$ in normalized MSE. Changing the number of terms preserves the theoretical MSE $p_4(B)$ but produces nonmonotone changes in exceedance frequency at the lowest threshold. Higher-threshold events are infrequent or unobserved in several conditions, showing that coefficient concentration affects tail behavior even when the second moment is fixed.

\subsection{Sign Risk Across Target Magnitudes}
\label{app:target_magnitude_sensitivity}

\paragraph{Experimental setup.}
We compare
\[
g_{\mathrm{sparse}}(z)=z_3z_4,
\qquad
g_{\mathrm{dense}}(z)=K^{-1/2}\sum_{3\leq a<b\leq d}z_az_b,
\qquad
K=\binom{d-2}{2}.
\]
Both give exact target MSE $\lambda^2p_4(B)$. We vary the positive ratio $\mu/\lambda$ from $0.02$ to $1.20$, including discrete error boundaries and the Experiment~2 reference ratio $0.8$. Strict sign reversal is the event $\widehat\theta_S<0$, and zero estimates are counted separately using exact lattice comparisons. We compare MSE, reversal and tie probabilities, and the paired dense-minus-sparse sign-risk difference with their exact finite-distribution references.

\begin{figure}[ht]
    \centering
    \includegraphics[width=\linewidth]{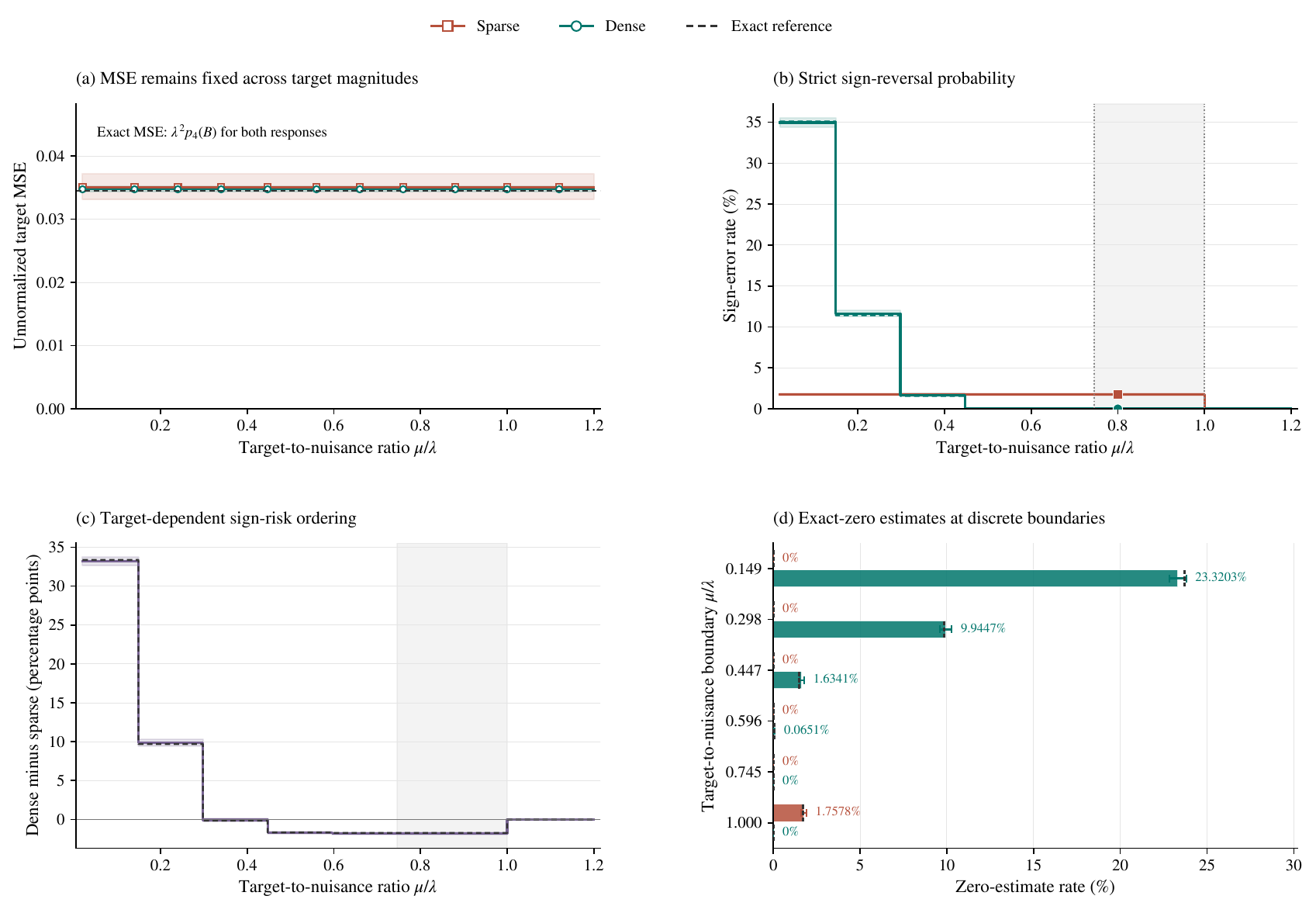}
    \caption{Target-magnitude sensitivity: (a) MSE, (b) strict sign-reversal probability, (c) dense-minus-sparse sign risk in percentage points, and (d) zero-estimate probabilities at discrete boundaries. The shaded interval $[5/\sqrt{45},1)$ gives zero dense reversal risk and positive sparse reversal risk.}
    \label{fig:app_target_magnitude}
\end{figure}

\paragraph{Results and analysis.}
Figure~\ref{fig:app_target_magnitude} shows unchanged MSE but a reversal in the ordering of sign risks as the target grows. Dense nuisance produces more sign reversals for weak targets, then declines in steps below the sparse risk. Sparse reversal probability is $p_4(B)/2$ for $0<\mu/\lambda<1$ and zero for $\mu/\lambda\ge1$. Dense strict reversal is impossible once $\mu/\lambda\ge5/\sqrt{45}$. Hence the interval $[5/\sqrt{45},1)$ gives strict sign-risk separation. At its left endpoint, dense estimates can still equal zero; at $\mu/\lambda=1$, sparse ties have probability $p_4(B)/2$. The tie probabilities distinguish avoiding a reversed sign from recovering a strictly positive estimate.

\subsection{Sensitivity to Dimension and Query Budget}
\label{app:dimension_budget_sensitivity}

\paragraph{Experimental setup.}
We use
\[
h_{d,k}(z)
=
z_1z_2+
\binom{d-2}{k}^{-1/2}
\sum_{\substack{R\subseteq\{3,\ldots,d\}\\|R|=k}}\phi_R(z),
\qquad
k\in\{2,4,6\}.
\]
These responses have target coefficient one, nuisance and surviving energies one, and response variance two. Every nuisance term is disjoint from the target, so exact MSE is $p_{k+2}(B)$. We cross $d\in\{8,15,31,63,127\}$ with $B\in\{32,64,128,256,512\}$, retaining $d<B/2$ and $B<2^d$. The budget sweep fixes $d=31$, and the dimension sweep fixes $B=256$. We also examine observed-to-exact MSE ratios for the sixth-order family across the retained grid and exceedance probabilities at twice the exact RMSE.

\begin{figure}[ht]
    \centering
    \includegraphics[width=\linewidth]{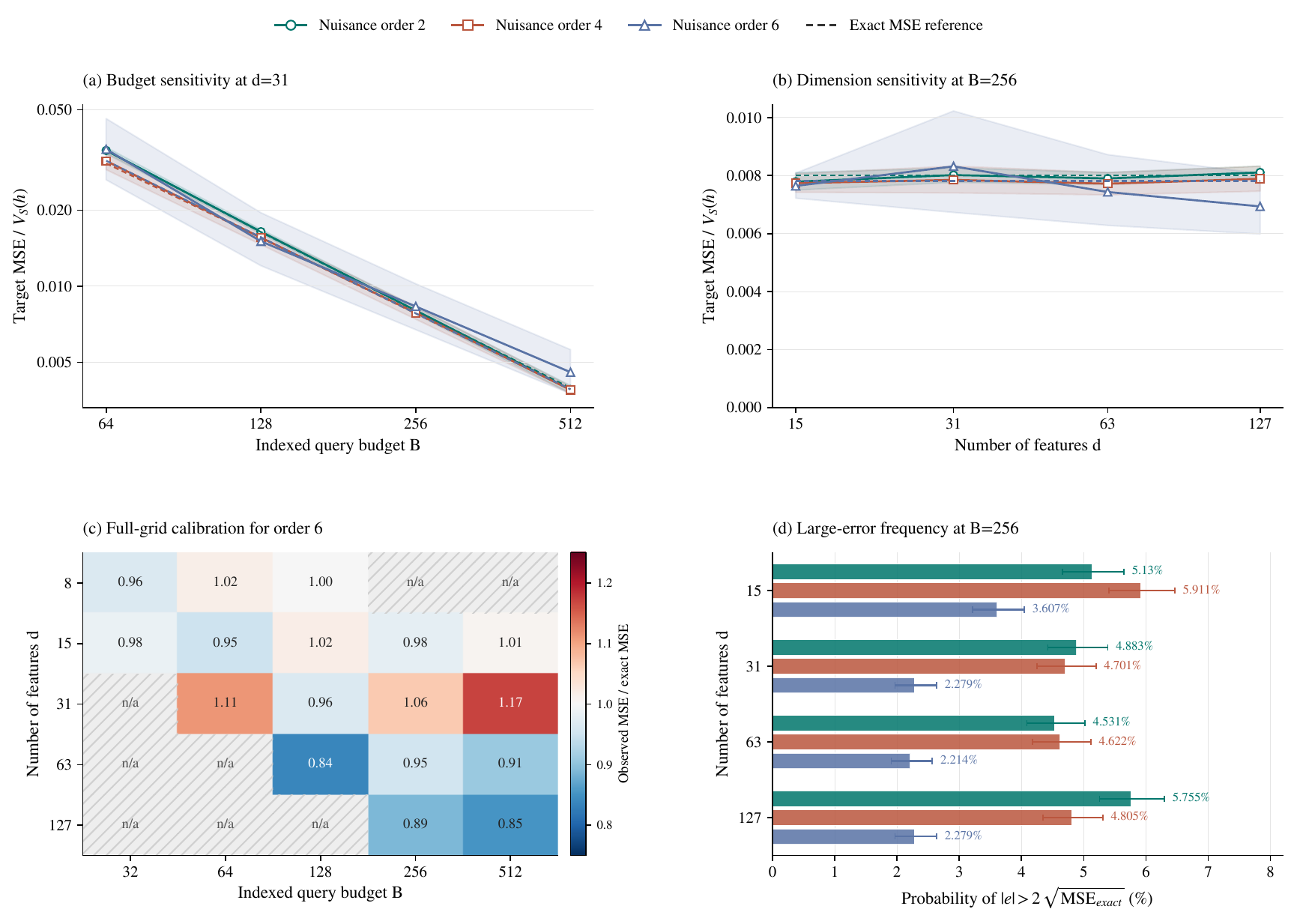}
    \caption{Dimension and budget sensitivity at fixed surviving energy: normalized MSE versus (a) budget at $d=31$ and (b) dimension at $B=256$; (c) observed-to-exact MSE ratios for sixth-order nuisance, with excluded cells hatched; (d) probabilities of absolute error exceeding twice the exact RMSE at $B=256$.}
    \label{fig:app_dimension_budget}
\end{figure}

\paragraph{Results and analysis.}
Figure~\ref{fig:app_dimension_budget} shows approximately inverse-budget MSE decay and estimates near the dimension-independent references at fixed budget. This dimension invariance follows from fixing both surviving energy and nuisance order relative to the target. The sixth-order family has wider intervals and more variable observed-to-exact ratios, including appreciable deviations from one in some cells. Exceedance frequencies at twice the exact RMSE show no monotone trend with dimension; the sixth-order family has fewer such events than the other two throughout the displayed dimension sweep.

\subsection{Sensitivity to Backgrounds and Evaluation Cases}
\label{app:background_case_sensitivity}

\paragraph{Experimental setup.}
We evaluate XGBoost and MLP on California Housing, white Wine Quality, and Adult using the preprocessing and training configurations of Experiment~4. Each model is fitted once and held fixed across three distinct training-row backgrounds and twelve held-out cases. Six cases come from the Experiment~4 evaluation pool, and six are selected outside that entire pool. Each background-case combination defines its own probe map, exact coefficients, and response variance. Original categorical variables switch as complete variables. CUBE and centered paired Monte Carlo each use $B=64$ query rows shared across all pair estimates.

Within each dataset--model group, we average all-pair normalized MSE equally over nonconstant probes in the relevant condition, then weight eligible groups equally. The same group set is used across backgrounds and case cohorts. Sign rates use normalized effect strength $|\theta_{ij}|/\mathrm{SD}(h)$, with bins $(0,0.01]$, $(0.01,0.05]$, and $(0.05,\infty)$. Numerically zero targets are excluded, and estimates within the numerical zero tolerance are recorded as ties. Within each bin, rates average eligible pair-probe combinations within groups and then give equal weight to groups represented in that bin.

\begin{figure}[ht]
    \centering
    \includegraphics[width=\linewidth]{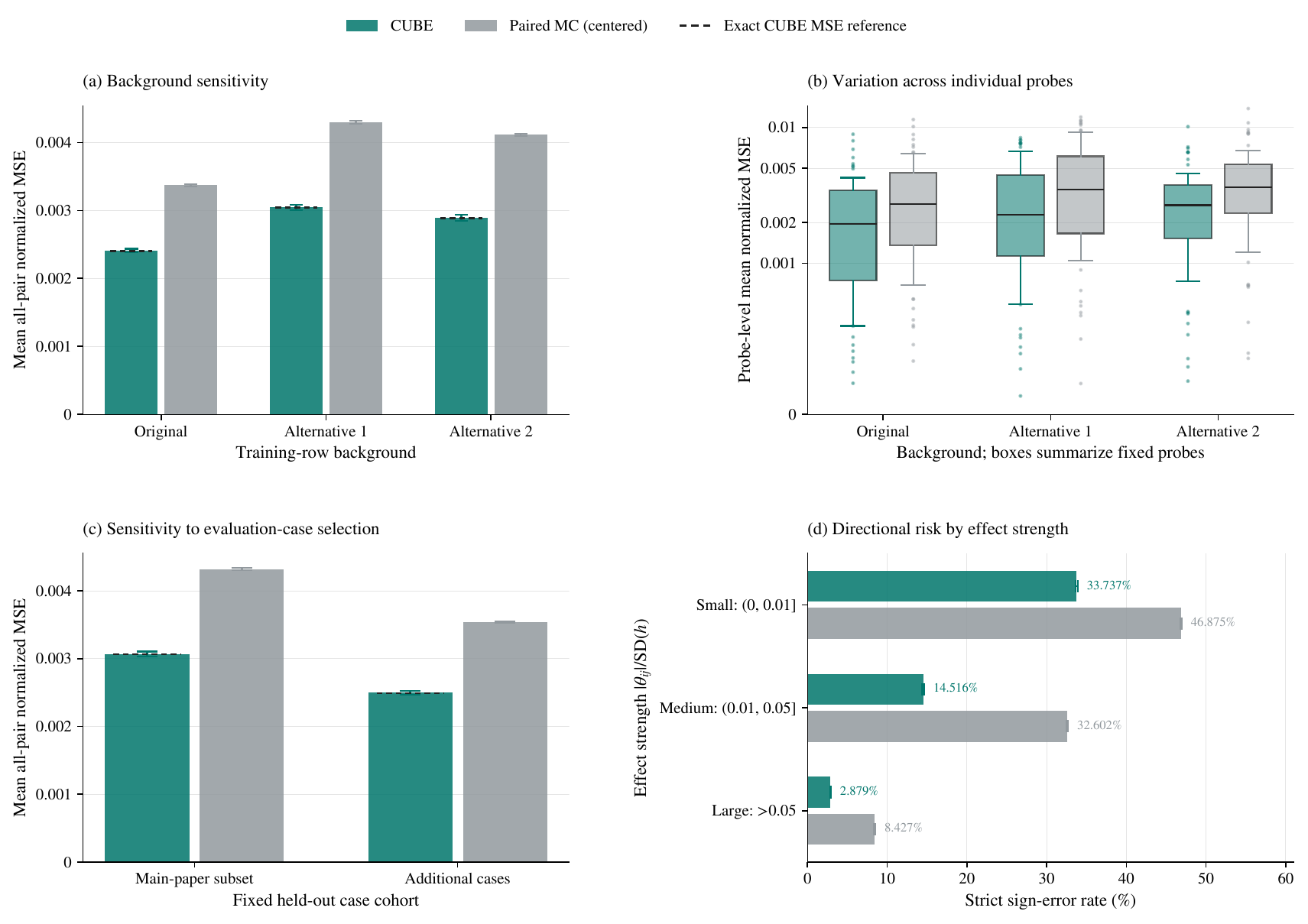}
    \caption{Background and case sensitivity for fixed models. Panels (a,c) show macro-averaged normalized MSE by background and case cohort, with exact CUBE references. Panel (b) shows probe-level normalized MSE on a symmetric-logarithmic axis; boxes span quartiles, central lines mark medians, and whiskers mark the $10$th and $90$th percentiles. Panel (d) shows sign-reversal rates by normalized effect strength, with numerical ties counted separately.}
    \label{fig:app_background_cases}
\end{figure}

\paragraph{Results and analysis.}
Figure~\ref{fig:app_background_cases} shows lower macro-averaged normalized MSE for CUBE across the three backgrounds, with estimates close to its aggregated exact references. Median probe-level errors are also lower, while the two methods' distributions overlap. The additional-case cohort has lower average error for both methods than the selected main-paper cohort, with a similar relative CUBE advantage. Sign-reversal rates decrease with effect strength and are lower for CUBE in each bin, although the weakest interactions retain substantial directional uncertainty.

\section{Reproducibility Details}
\label{app:reproducibility}

\paragraph{Data and preprocessing.}
\begin{itemize}
    \item \textbf{Sources and dimensions:} California Housing is loaded through \texttt{scikit-learn}; white Wine Quality and Adult use the UCI files. The probe dimensions are $8$, $11$, and $14$ original features, respectively.
    \item \textbf{Splits:} California Housing and Wine Quality use seeded $80/20$ train--test splits. Identical white-wine records are removed before splitting. Adult retains its official training and test partitions.
    \item \textbf{Adult conventions:} Question marks are treated as missing values, trailing periods are removed from test labels, and income above \$50K is the positive class. All fourteen original predictors are retained.
    \item \textbf{Numerical features:} Missing values are filled with training medians, followed by standardization using training statistics.
    \item \textbf{Categorical features:} Missing values receive the label \texttt{Missing}. One-hot encoding is fitted on training data with unseen categories ignored. Probes switch original variables before encoding, so their encoded columns change jointly.
    \item \textbf{Probe construction:} Each feature switches between its explained value and its value in a training-row background. Features with identical states are retained. Background and held-out case selections are shared across models within each dataset and experiment.
    \item \textbf{Responses:} California and Wine use regression predictions; Adult uses positive-class probabilities. Experiment~3 additionally constructs synthetic responses by permuting fourth-order coefficients.
\end{itemize}

\paragraph{Predictive models.}
\begin{itemize}
    \item \textbf{XGBoost:} $400$ trees; maximum depth $6$; learning rate $0.05$; row subsampling and column subsampling per tree both $0.9$; histogram-based construction. Objectives are squared error for regression and binary logistic loss for classification.
    \item \textbf{MLP architecture:} Hidden widths $(128,64)$; ReLU activations; regularization parameter $\alpha=10^{-4}$.
    \item \textbf{MLP training:} Batch size $256$; initial learning rate $10^{-3}$; maximum $400$ iterations; early stopping with validation fraction $0.15$ and \texttt{n\_iter\_no\_change}$=25$.
    \item \textbf{Parallelism:} XGBoost uses \texttt{n\_jobs}$=2$. Experiment~4 and the background-sensitivity experiment additionally limit supported numerical-library thread pools to two threads.
    \item \textbf{Fixed models:} Each fitted model remains fixed across explanation randomizations. The background-sensitivity experiment also holds each model fixed across all backgrounds and case cohorts.
\end{itemize}

\paragraph{Comparison implementations.}
\begin{itemize}
    \item \textbf{Package and scale:} Literature methods use \texttt{shapiq} version \texttt{1.7.0}. Returned pairwise Banzhaf or order-two Faith-Banzhaf interactions are divided by four to obtain the coefficient scale.
    \item \textbf{CUBE and paired Monte Carlo:} CUBE samples distinct nonzero labels and independent signs. Centered paired Monte Carlo uses $m=B/2$ independent uniform configurations and their reversals, with sample-covariance denominator $m-1$.
    \item \textbf{SHAP-IQ and SVARM-IQ:} Both use \texttt{index="BII"}, \texttt{max\_order=2}, \texttt{pairing\_trick=True}, and \texttt{top\_order=False}.
    \item \textbf{Faith-Banzhaf-2:} The \texttt{RegressionFBII} implementation uses \texttt{max\_order=2} and paired sampling. Its coalition-size sampling weights are $\binom{d}{k}/2^d$ for $k=0,\ldots,d$.
    \item \textbf{ProxySPEX:} Settings are \texttt{index="FBII"}, \texttt{max\_order=2}, \texttt{pairing\_trick=False}, and \texttt{top\_order=False}. The surrogate is an XGBoost regressor with $100$ trees, learning rate $0.1$, full row and column sampling, \texttt{reg\_lambda=1}, and \texttt{min\_child\_weight=1}.
    \item \textbf{Surrogate selection:} Depth candidates are $\{2,4,6\}$, evaluated by shuffled three-fold cross-validation using prediction MSE. The selected surrogate is refitted on all queried responses. Training and validation use the same queried pool.
    \item \textbf{Query accounting:} Budgets count requested rows, including repetitions. Cached responses are charged in the same way as direct evaluations, and repeated inputs do not free budget for replacement queries.
\end{itemize}

\paragraph{Randomization.}
\begin{itemize}
    \item \textbf{Root seeds:} Experiments~1--4, target-magnitude sensitivity, and background sensitivity use $20260906$. Design ablation uses $20260912$; nuisance-structure sensitivity uses $20260913$; dimension-budget sensitivity uses $20260915$.
    \item \textbf{Support selection:} Nuisance-structure supports and coefficient signs use the separate seed $20260914$ and remain fixed across design randomizations.
    \item \textbf{Dataset and model identifiers:} Dataset identifiers are $0$, $1$, and $2$ for California, Wine, and Adult; model identifiers are $0$ and $1$ for XGBoost and MLP.
    \item \textbf{Splitting and fitting:} For root seed $R$, dataset identifier $a$, and model identifier $b$, randomized train--test splits use $R+a$, and predictive-model fitting uses $R+10000a+b$.
    \item \textbf{Sampling streams:} NumPy generators use spawned or explicitly tagged \texttt{SeedSequence} streams. Stream tags separate background selection, case selection, estimator randomization, and bootstrap resampling, while preserving the pairing specified for each experiment.
\end{itemize}

\paragraph{Numerical conventions.}
\begin{itemize}
    \item \textbf{Precision:} Floating-point contrasts, coefficient spectra, and error summaries use double precision.
    \item \textbf{Target eligibility:} Real-model sign scoring excludes targets with $|\theta_{ij}|\leq\tau(h)$, where $\tau(h)=64\epsilon\max_z|h(z)|$ and $\epsilon$ is double-precision machine epsilon.
    \item \textbf{Experiment~4 sign diagnostic:} The saved sign-error statistic compares $\operatorname{sign}(\widehat\theta_{ij})$ with $\operatorname{sign}(\theta_{ij})$ for eligible targets. An exactly zero estimate therefore counts as a sign mismatch.
    \item \textbf{Background-sensitivity sign diagnostic:} Estimates with $|\widehat\theta_{ij}|\leq\tau(h)$ are recorded as ties. A sign reversal requires an opposite-sign estimate outside this tolerance.
    \item \textbf{Discrete boundaries:} Target-magnitude sensitivity uses exact lattice comparisons for reversals and ties. Dimension-budget exceedances use integer contrasts and exact threshold comparisons.
    \item \textbf{Constant probes:} Experiment~3 retains constant-response cases with zero error. Experiment~4 and background sensitivity exclude constant probes from variance-normalized averages and retain their raw losses.
\end{itemize}

\paragraph{Execution and saved outputs.}
\begin{itemize}
    \item \textbf{Execution:} Each experiment is organized as a Google Colab cell with its configuration at the top.
    \item \textbf{Runtime requirements:} Experiment~4 requires Python~3.12 or later and \texttt{shapiq==1.7.0}. Use a fresh runtime if a different \texttt{shapiq} version has already been imported. Other dependency versions are recorded in the experiment outputs.
    \item \textbf{Records:} Outputs include settings, seeds, numerical summaries, randomization results, and runtime package versions. Real-model records also include data selections, fitted-model parameters, predictive performance, and training warnings.
    \item \textbf{Benchmark diagnostics:} Experiment~4 records query counts, estimator warnings, and sampled quadratic-design ranks for Faith-Banzhaf-2.
    \item \textbf{Artifacts and caches:} The background-sensitivity archive includes fitted models and response cubes under the default settings. Compatible caches allow repeated analyses with the same fitted models and probe responses.
\end{itemize}

\end{document}